\documentclass[sigconf]{acmart}

\copyrightyear{2026}
\acmYear{2026}
\setcopyright{cc}
\setcctype{by}
\acmConference[KDD '26]{Proceedings of the 32nd ACM SIGKDD Conference on Knowledge Discovery and Data Mining V.2}{August 09--13, 2026}{Jeju Island, Republic of Korea}
\acmBooktitle{Proceedings of the 32nd ACM SIGKDD Conference on Knowledge Discovery and Data Mining V.2 (KDD '26), August 09--13, 2026, Jeju Island, Republic of Korea}
\acmDOI{10.1145/3770855.3819010}
\acmISBN{979-8-4007-2259-2/2026/08}

\usepackage{booktabs}
\usepackage{siunitx}
\usepackage{multirow, makecell}

\title{Explainable AI for Cancer Drug Response Prediction: Beyond Univariate Feature Attributions}

\author{Martino Ciaperoni}
\authornote{Equal contribution.}
\orcid{0009-0009-7581-2031}
\affiliation{
  \institution{\href{https://kdd.isti.cnr.it/}{KDD Lab}, Scuola Normale Superiore}
   \city{Pisa}
  \country{Italy}
}
\email{martino.ciaperoni@sns.it}

\author{Margherita Lalli}
\authornotemark[1]
\orcid{0009-0001-0653-0278}
\affiliation{
  \institution{\href{https://kdd.isti.cnr.it/}{KDD Lab}, Scuola Normale Superiore}
   \city{Pisa}
  \country{Italy}
}
\email{margherita.lalli@sns.it}

\author{Simone Piaggesi}
\authornotemark[1]
\orcid{0000-0001-9710-6255}
\affiliation{
  \institution{\href{https://kdd.isti.cnr.it/}{KDD Lab}, University of Pisa}
   \city{Pisa}
  \country{Italy}
}
\email{simone.piaggesi@di.unipi.it}

\author{Martina Varisco}
\orcid{0000-0001-9703-7798}
\affiliation{
  \institution{Bio@SNS, Scuola Normale Superiore}
   \city{Pisa}
  \country{Italy}
}
\email{martina.varisco@sns.it}

\author{Francesco Carli}
\orcid{0009-0006-6980-9440}
\affiliation{
  \institution{EMBL-EBI}
  \city{Hinxton}
  \country{United Kingdom}
}
\email{carli@ebi.ac.uk}

\author{Riccardo Guidotti}
\orcid{0000-0002-2827-7613}
\affiliation{
  \institution{\href{https://kdd.isti.cnr.it/}{KDD Lab}, University of Pisa, and ISTI-CNR}
  \city{Pisa}
  \country{Italy}
}
\email{riccardo.guidotti@unipi.it}

\author{Dino Pedreschi}
\orcid{0000-0003-4801-3225}
\affiliation{
  \institution{\href{https://kdd.isti.cnr.it/}{KDD Lab}, University of Pisa}
  \city{Pisa}
  \country{Italy}
}
\email{dino.pedreschi@unipi.it}

\author{Francesco Raimondi}
\orcid{0000-0002-6891-3178}
\affiliation{
  \institution{Bio@SNS, Scuola Normale Superiore}
  \city{Pisa}
  \country{Italy}
}
\email{francesco.raimondi@sns.it}

\author{Fosca Giannotti}
\orcid{0000-0003-3099-3835}
\affiliation{%
  \institution{\href{https://kdd.isti.cnr.it/}{KDD Lab}, Scuola Normale Superiore}
  \city{Pisa}
  \country{Italy}
}
\email{fosca.giannotti@sns.it}

\usepackage{xspace}
\usepackage{tikz}
\usetikzlibrary{chains,arrows.meta,positioning,calc}
\usepackage{fontawesome5} 
\usepackage{xcolor}
\usepackage{amsmath}
\usepackage{tcolorbox}
\usepackage{subcaption}

\definecolor{pathcol}{RGB}{180,30,30}
\definecolor{boxbg}{RGB}{248,248,248}

\tcbset{
  rulebox/.style={
    colback=boxbg,
    colframe=black,
    boxrule=0.6pt,
    arc=2pt,
    left=6pt,
    right=6pt,
    top=6pt,
    bottom=6pt
  }
}

\newcommand{\spara}[1]{\vspace{0.5em}\noindent\textbf{#1}}

\newcommand{\xvec}{\mathbf{x}}

\newcommand{\graphPol}{\mathcal{G}_{\Delta}\xspace}

\newcommand{\drug}[1]{\textsc{#1}}

\newcommand{\ILLUME}{\textsc{ILLUME}\xspace}
\newcommand{\BORUTA}{\textsc{Boruta}\xspace}

\newcommand{\sensitive}{sensitive\xspace}
\newcommand{\resistant}{resistant\xspace}

\newcommand{\illora}{ILLUME+\xspace}
\newcommand{\ILLoRA}{\illora}

\usepackage{hyperref}
\usepackage{cleveref}
\usepackage{enumitem}

\usepackage{makecell}

\usepackage{xcolor}

\newcommand{\dataset}{\mathcal{D}}
\newcommand{\drugdataset}[1]{\dataset^{(#1)}}
\newcommand{\feat}{\mathbf{x}}
\newcommand{\labely}{y}
\newcommand{\model}[1]{f^{(#1)}}
\newcommand{\explainer}{\ensuremath{\mathcal{E}}\xspace}
\newcommand{\explanation}{e}

\newcommand{\pg}[1]{\textcolor{pathcol}{\texttt{#1}}}
\newcommand{\gene}[1]{\texttt{#1}}

\usepackage{booktabs}

\begin{document}

\begin{abstract}
{
Predicting cancer drug response from transcriptomic profiles is a cornerstone of precision oncology, yet the scientific value of machine learning models hinges not solely on predictive accuracy, but also on their capacity to generate reliable biological insights. 
Current explainability approaches in this setting are computationally costly, lack robustness, and reduce complex drug response to univariate gene importance scores, overlooking the coordinated gene activity that drives sensitivity and resistance.

In this work, we present ILLUME+, a scalable post-hoc explainability framework that moves beyond single-gene assessments to capture multiple, complementary forms of explanation. Integrated into 
our
end-to-end pipeline, ILLUME+ produces more stable gene importance scores than existing baselines, recovers established drug-gene associations and mechanisms of action, and enables AI-assisted hypothesis generation to uncover novel interaction-driven molecular signals in cancer biology.
}
\end{abstract}



\maketitle

\section{Introduction}\label{sec:introduction} 
Cancer remains a major cause of mortality worldwide, and the substantial variability in response to anticancer therapies due to  molecular heterogeneity makes treatment selection a persistent challenge~\cite{sung2021global}.
The growing availability of large-scale pharmacogenomic resources, such as the \emph{Genomics of Drug Sensitivity in Cancer} database (GDSC)~\cite{iorio2016landscape}, enables systematic investigation of the relationship between tumor molecular profiles and drug response. In this context, machine learning has become a central analytical tool, allowing the modeling of high-dimensional, nonlinear dependencies and achieving strong predictive performance across diverse cancer cell lines~\cite{chen2021survey, kuenzi2020predicting}.

However, predictive performance alone is insufficient to drive scientific progress.
To inform biological understanding, models must also be amenable to explanation, 
enabling the identification of molecular mechanisms underlying drug sensitivity and resistance~\cite{rees2016correlating}.
Approaches that embed prior biological knowledge into models aim to address this limitation~\cite{tang2021explainable,shi2025drexplainer}, but may bias discovery toward established mechanisms and limit genuinely data-driven insights~\cite{samal2022opportunities}.
It remains an open question whether high-performing drug response models can uncover biologically meaningful structure without relying on predefined assumptions.

Explainable Artificial Intelligence (XAI) provides a natural framework for addressing this challenge~\cite{guidotti2018survey}. Yet, existing XAI studies in drug response prediction rely primarily on univariate feature attribution methods, particularly SHAP~\cite{lundberg2017unified} which, however, suffer from two major limitations. First, SHAP-based explanations exhibit limited robustness and scalability to high-dimensional settings~\cite{piaggesi2025explanations}. 
More fundamentally, univariate explanations assess genes independently, whereas cellular response to treatment arises from coordinated activity among multiple genes and pathways. 
As a result, interactions and higher-order biological mechanisms contributing to drug response remain hidden. 

To address this, we build on \textsc{ILLUME}~\cite{piaggesi2025explanations}, a post-hoc XAI framework that learns faithful local approximations of black-box predictors and supports multiple explanation modalities — feature attributions, decision rules, and counterfactual explanations — better reflecting the multivariate nature of biological systems. Yet its direct application to transcriptomic data is impractical, as the computational cost scales poorly with input dimensionality. 
We therefore introduce \illora, an extension of \textsc{ILLUME} designed for high-dimensional data that retains its rich explanatory objects capturing more-than-univariate gene signals.

\begin{figure}[t]
    \centering
    \includegraphics[trim={6cm, 0, 4cm, 0}, clip, width=0.8\linewidth]{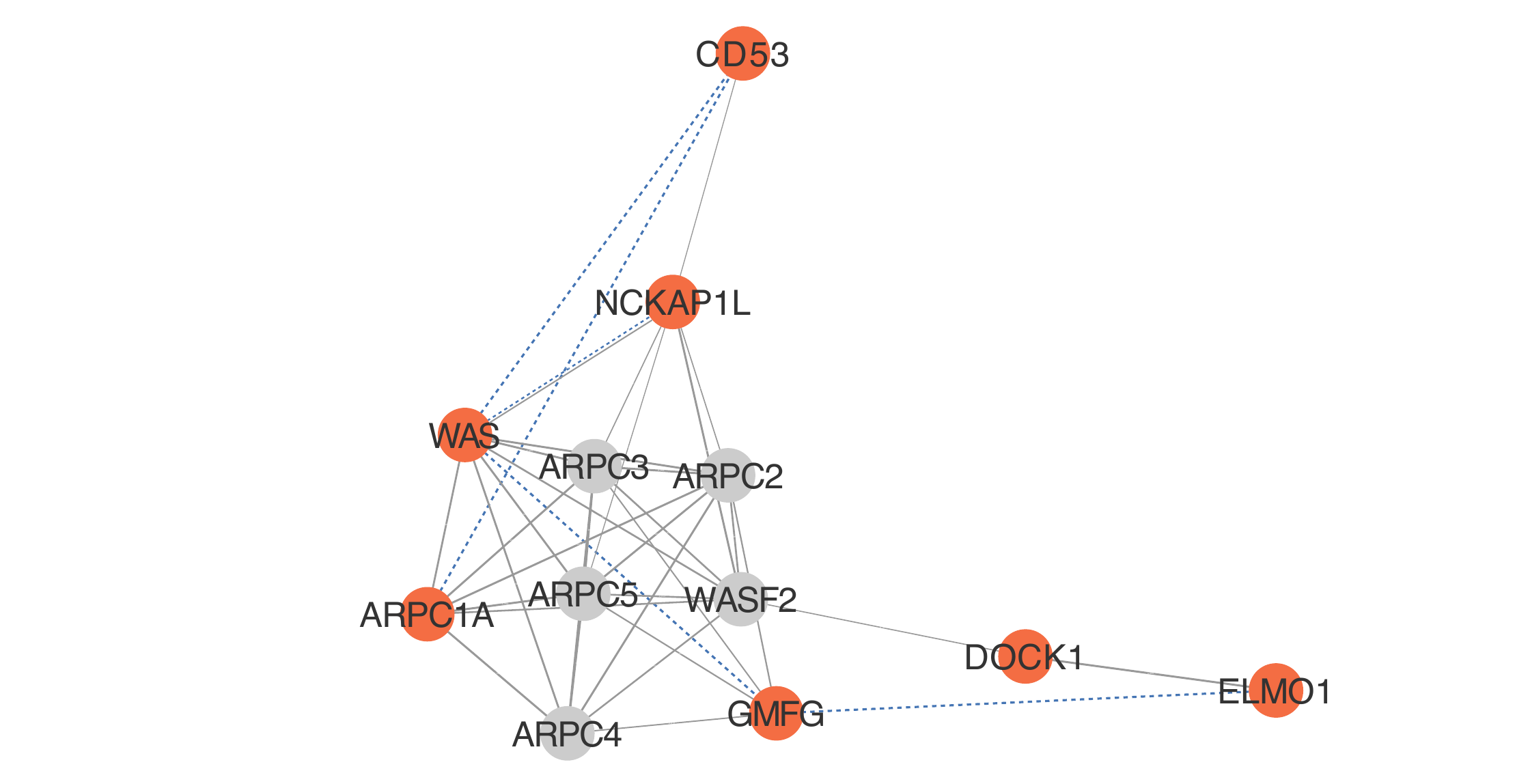}
    \caption{
    Example gene graph for \drug{Venetoclax}-sensitive cell lines. Strongly interacting genes identified by our approach are in red, neighboring genes from  the STRING database in grey. Solid edges indicate high-confidence interactions from STRING, dashed  links are inferred by our approach.
    }
    \label{fig:selling_figure}
\end{figure}

Figure~\ref{fig:selling_figure} shows an example of pairwise gene-gene interactions retrieved by \illora. With no biological prior, it can partially recapitulate high-confidence interaction graphs from STRING, a resource for functional protein networks integrating curated databases, experiments, co-expression, gene context and text mining \cite{szklarczyk2023string}. \illora captures long-range signaling coordination in immunity and cell shape remodeling~\cite{rohatgi1999interaction}, suggesting potential new gene connections.
Moreover, four of the five missing STRING nodes result from prior feature selection rather than limitations of \illora, while the remaining node was absent from the original database.

\spara{Contributions.}
Our contributions can be summarized as follows:
\begin{itemize}[leftmargin=*]
    \item We introduce \illora, a scalable extension of \textsc{ILLUME} that enables the extraction of robust, structured explanations from high-dimensional  data, addressing key scalability limitations of existing post-hoc XAI methods.

    \item We design a 
    fully data-driven, end-to-end 
    pipeline based on \illora\ to move beyond univariate feature attributions and recover 
    explanations that capture gene-gene mechanisms underlying drug sensitivity and resistance.

    \item  We demonstrate how our method can confirm known biological associations and reveal gene interactions to generate testable hypotheses that can support AI-assisted biological discovery without relying on prior knowledge.

\end{itemize}
We made our source code and artifacts publicly available at:\\ \url{https://github.com/simonepiaggesi/illume-plus/}.

\spara{Roadmap.} The paper is organized as follows. Section~\ref{sec:related} reviews related work on explainability methods and drug sensitivity prediction.  
Section~\ref{sec:setting} introduces the preliminaries, and Section~\ref{sec:method} describes our methodology. 
In Section~\ref{sec:results} and Section~\ref{sec:discussion} we present quantitative and qualitative results and further discuss them.

\section{Related Work} 
\label{sec:related}
\spara{Explainable AI for science.}
The growing emphasis on Explainable Artificial Intelligence (XAI) reflects its emerging role as a prerequisite for scientific discovery across disciplines, including healthcare~\cite{XAIhealth}, material science~\cite{XAImaterial}, climatology~\cite{XAIclimate}, and biology~\cite{XAIBio}. 
Beyond improving model transparency, XAI has enabled mechanistic insights from complex predictive systems, particularly in biomedical settings. For instance, tools like DeepSHAP have been successfully applied to  AlphaFold2 models to understand their reasoning and detect the role of specific amino acids in different protein structures~\cite {sibli2025enhancing}. 
In transcriptomics and drug discovery, SHAP-based approaches have been widely employed to identify relevant genes as well as predicting synergies between drugs and between drugs and cell lines~\cite{janizek2023synergistic,carli2025learning,keyl2025neural}. However, these methods are typically restricted to univariate explanations and, in high-dimensional settings, often rely on 
model-specific implementations tailored to 
tree-based predictors (e.g., TreeSHAP~\cite{lundberg2020local}).

\spara{Drug response prediction.}
Drug response prediction has evolved from linear and sparsity-inducing models~\cite{iorio2016landscape,rees2016correlating}, interpretable but poorly accurate, toward more expressive nonlinear approaches, including gradient-boosting and deep neural networks~\cite{kuenzi2020predicting,chen2021survey}. To improve interpretability,  biological priors such as pathways, gene networks, or drug targets have been incorporated into the model architecture~\cite{tang2021explainable,shi2025drexplainer,carli2025learning}.
While these approaches can improve both performance and interpretability, they also shape the conclusions that can be drawn from the data, making it difficult to disentangle learned biological structure from assumptions imposed a priori, and failing to provide an unbiased view of the spectrum of associations and patterns present in the data~\cite{samal2022opportunities}.

\section{Preliminaries}\label{sec:setting}
This section introduces the setting needed to understand our contributions at the intersection of transcriptomics and XAI.

\spara{Pharmacogenomic and transcriptomic context.}
Effective cancer treatment requires drugs that selectively target molecular vulnerabilities in the patient's cancer cells. Large-scale pharmacogenomic resources have become central to this effort by systematically measuring the relationship between tumor molecular profiles and drug response. We use transcriptomic data from the \emph{Cancer Cell Line Encyclopedia} (CCLE)~\cite{barretina2012cancer} to learn drug response from GDSC database.
Drugs exert their action by interacting with specific proteins, which are encoded by their corresponding genes; we refer to these genes as \textit{putative targets}.
Their biochemical interaction with the drug and its effect define the \textit{mechanism-of-action} (MoA) of the drug. 
However, drug response is not determined solely by the presence or activity of individual targets or MoA-related genes. Instead, it emerges from the broader cellular state.
This global context can be approximated through the activity of \textit{gene pathways}, i.e., sets of genes that act in a coordinated manner to carry out shared cellular processes.
Gene enrichment analysis~\cite{subramanian2005gene} has traditionally been used to assess whether selected gene sets show overrepresentation of known pathways or functional categories, thereby linking predictive signals to established cellular processes and facilitating mechanistic insights. However, this technique is limited to projecting genes in known pathways, and it does not allow the discovery of new gene relationships.
In this study, we validate our approach at three levels of drug action previously mentioned: recognition of putative targets, MoA, and global pathway-level effects.

\spara{Explainable AI.}
In this work, we consider tabular datasets defined as
$
\mathcal{D} = \{(\feat_i, \labely_i)\}_{i=1}^{N}, 
$
where $\feat_i \in \mathcal{X}$ denotes the feature set of the $i$-th instance, belonging to the feature space $\mathcal{X} \subseteq \mathbb{R}^{m}$, and $\labely_i \in \mathcal{Y}$ is the label to be predicted.
Data instances represent cell lines, and features correspond to expression measurements of genes $g_1, \dots g_m$. 
Given a trained predictive model $f:\mathcal{X} \rightarrow \mathcal{Y}$ learned from $\mathcal{D}$, we broadly define an explainer as a mapping:
$
\explainer : \big(f, \feat \big) \;\mapsto\; \explanation(\feat),
$
where $\explanation(\feat)$ denotes a local explanatory object describing the model’s behavior in the neighborhood of $\feat$.
Depending on the explainer, $\explanation(\feat)$ may take different forms, including feature attribution vectors, local surrogate models, logical decision rules, or higher-order structures capturing dependencies among input features. 
\\
Our approach to constructing 
$\explanation(\feat)
$ builds upon ILLUME~\cite{piaggesi2025explanations}, which we adopt as a starting point and extend. 
ILLUME is a post-hoc explanation framework recently introduced to extract robust, instance-level explanations for black-box predictive models. Given an input $\xvec \in \mathbb{R}^m$ and a trained black-box model $f$, \ILLUME learns a \emph{meta-encoder} that maps each input to a low-dimensional latent representation $
\mathbf{z} \in \mathbb{R}^v
$ while preserving the local decision structure of the black-box.
Specifically, \ILLUME learns an instance-specific linear map $W(\mathbf{x})\in\mathbb{R}^{v\times m}$ with a multi-layered hypernetwork~\cite{HaDL17}. This locally-linear projection transforms input instances into compact latent representations 
$
\mathbf{z} = W(\mathbf{x})\mathbf{x},
$
where each row of $W(\mathbf{x})$ is normalized to unit $\ell_2$ norm. 
The latent space is optimized using similarity-preserving KL divergence losses, alongside regularization promoting orthogonality, decorrelation, and smoothness via a Jacobian penalty. 
This design makes \ILLUME a \emph{meta-explainer}: a global \ILLUME is trained once, yet it can generate instance-specific explanations on demand by fitting an interpretable surrogate (e.g., logistic regression or a shallow decision tree) in the locally-linear latent space. Explanations are obtained by decoding surrogate decisions back to the input space through $W(\mathbf{x})$, yielding, e.g., feature attributions, decision rules, and counterfactual explanations. 

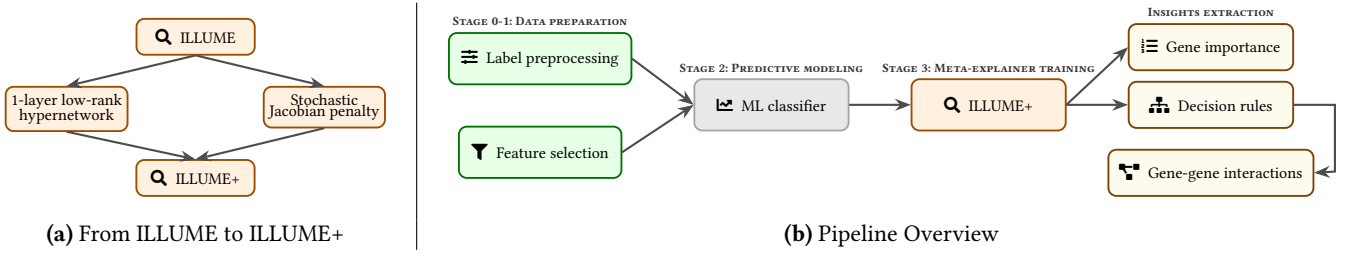
\begin{figure*}[t]
\centering

\begin{tabular}{c@{\hspace{4mm}}|@{\hspace{4mm}}c}


\begin{tikzpicture}[
    scale=0.68,
    transform shape,
    explainfill/.style={fill=orange!12, draw=orange!60!black},
    stage/.style={
        rounded corners=3pt,
        align=center,
        inner xsep=3pt,
        inner ysep=4pt,
        minimum width=2.3cm,
        minimum height=7mm,
        line width=0.7pt
    },
    arrow/.style={
        -{Stealth[length=2mm]},
        line width=0.8pt,
        draw=black!70
    }
]

\node[stage, explainfill] (illume) {\faSearch\; ILLUME};

\node[
    stage,
    explainfill,
    below left=6mm and 1.5mm of illume,
    minimum width=2.1cm,
] (lowrank) {1-layer low-rank\\hypernetwork};

\node[
    stage,
    explainfill,
    below right=6mm and 1.5mm of illume,
    minimum width=2.1cm,
] (jac) {Stochastic\\Jacobian penalty};

\coordinate (midmods) at ($(lowrank)!0.5!(jac)$);

\node[
    stage,
    explainfill,
    below=10mm of midmods
] (illumeplus) {\faSearch\; ILLUME+};

\draw[arrow] (illume.south) -- (lowrank.north);
\draw[arrow] (illume.south) -- (jac.north);
\draw[arrow] (lowrank.south) -- (illumeplus.north);
\draw[arrow] (jac.south) -- (illumeplus.north);

\end{tikzpicture}

&


\begin{tikzpicture}[
    scale=0.68,
    transform shape,
    prepfill/.style={fill=green!10, draw=green!45!black},
    blackboxfill/.style={fill=gray!18, draw=gray!70},
    explainfill/.style={fill=orange!12, draw=orange!60!black},
    outfill/.style={fill=yellow!75!gray!10, draw=orange!50!black},
    stage/.style={
        rounded corners=3pt,
        align=center,
        inner xsep=7pt,
        inner ysep=6pt,
        minimum width=3.0cm,
        minimum height=10mm,
        line width=0.7pt
    },
    output/.style={
        rounded corners=3pt,
        align=center,
        inner xsep=6pt,
        inner ysep=5pt,
        minimum width=3.2cm,
        minimum height=9mm,
        line width=0.7pt
    },
    arrow/.style={
        -{Stealth[length=2.1mm]},
        line width=0.8pt,
        draw=black!70
    }
]
    \node[stage, prepfill] (s1) {\faSlidersH\; Label preprocessing};
    \node[stage, prepfill, below=8mm of s1] (s2) {\faFilter\; Feature selection};
    
    \node[stage, blackboxfill, right=12mm of s1, yshift=-9mm] (s3) {\faChartLine\; ML classifier};
    \node[stage, explainfill, right=12mm of s3] (s4) {\faSearch\; \illora};
    
    \node[output, outfill, right=12mm of s4] (o2) {\faSitemap\; Decision rules};
    \node[output, outfill, above=2mm of o2] (o1) {\faListOl\; Gene importance};
    \node[output, outfill, below=4mm of o2] (o3) {\faProjectDiagram\; Gene-gene interactions};
    
    \draw[arrow] (s1.east) -- (s3.west);
    \draw[arrow] (s2.east) -- (s3.west);
    \draw[arrow] (s3.east) -- (s4.west);
    \draw[arrow] (s4.east) -- (o1.west);
    \draw[arrow] (s4.east) -- (o2.west);
    


   \draw[arrow]
    (o2.east)
    -- ++(8mm,0)
    -- ++(0,-13mm)
    -- (o3.east);
    
    \node[font=\footnotesize\bfseries\scshape, text=black!80, above=0.2mm of s1] {Stage 0-1: Data preparation};
    \node[font=\footnotesize\bfseries\scshape, text=black!80, above=0.2mm of s3] {Stage 2: Predictive modeling};
    \node[font=\footnotesize\bfseries\scshape, text=black!80, above=0.2mm of s4] {Stage 3: Meta-explainer training};
    \node[font=\footnotesize\bfseries\scshape, text=black!80, above=0.2mm of o1] {Insights extraction};
\end{tikzpicture}

\\[2mm]

\textbf{(a)} From ILLUME to ILLUME+
&
\textbf{(b)} Pipeline Overview

\end{tabular}

\caption{
(a) ILLUME+ extends ILLUME through a 1-layer low-rank hypernetwork and a stochastic Jacobian penalty, improving scalability. 
(b) Schematic overview of our pipeline. Stages 0-1 preprocess data to feed to the black-box predictive model (Stage 2), which outputs predictions that are fed to ILLUME+ (Stage 3). ILLUME+ produces different explanatory objects: gene attributions and decision rules, from which gene--gene relevance are extracted.
}
\label{fig:pipeline}

\end{figure*}

\section{Methodology}
\label{sec:method}
This section describes our methodology for identifying genes and gene 
groups
that drive drug response predictions.

\subsection{\illora: scalable post-hoc explanations}

As discussed in Sections~\ref{sec:introduction}~and~\ref{sec:setting}, while related work hinges upon SHAP for extracting explanations, our work identifies the ILLUME meta-explainer~\cite{piaggesi2025explanations} as a more promising method to extract explanation objects that are not limited to (unstable) univariate feature attributions. 
However, like SHAP-based approaches,  \ILLUME faces practical challenges when applied to complex domains characterized by high-dimensional tabular data, such as drug response prediction.

To handle the dimensionality typical of transcriptomic data while preserving the key  advantages of \ILLUME and remaining fully model-agnostic with respect to the black-box predictive model,  we introduce \ILLoRA, 
a scalable extension of \ILLUME designed to operate effectively on tabular datasets with large feature spaces. 
As depicted in Figure~\ref{fig:pipeline} (\textbf{a}),
\illora preserves the same objective and explanation semantics 
as \ILLUME 
while substantially reducing computational cost by targeting the two main scalability bottlenecks: 
the prohibitive amount of parameters of the hypernetwork and the Jacobian penalty computation.  
To 
reduce the parameters count of \ILLUME, \illora employs a simpler 1-layer hypernetwork~\cite{HaDL17} whose weights are compressed via low-rank factorization. 
In particular, \illora replaces the full hypernetwork used to predict $W(\mathbf{x}) \in \mathbb{R}^{v \times m}$ with a 1-layer low-rank decomposition of rank $r \ll m$. 
Rather than optimizing the full $(v \times m) \times m$ parameters of the 1-layer architecture, 
for each latent component $l \in \{1,\dots,v\}$,  we train low-rank weights $\{U_l \in \mathbb{R}^{r \times m}\}_{l=1\dots v}$, $B\in \mathbb{R}^{r \times m}$ and $\mathbf{b} \in \mathbb{R}^{m}$ such that 
\[
u_l(\mathbf{x}) = U_l \mathbf{x} \in \mathbb{R}^{r}, \qquad
w_l(\mathbf{x}) = B^\top\, u_l(\mathbf{x}) + \mathbf{b} \in \mathbb{R}^{m},
\]
and stack the resulting $w_l(\mathbf{x})$ to form $W(\mathbf{x})$. 
This parameterization dramatically reduces the order of the trainable parameters 
to $(v \times m) \times r$, allowing scaling to transcriptomic settings that involve thousands of genes.
Secondly, to tackle the computational cost of the Jacobian 
regularizer
while retaining the robustness it guarantees, 
\illora replaces the full Jacobian penalty with an efficient stochastic estimator based on Jacobian--vector products~\cite{hoffman2019robust}. 
The regularizer is applied to the residual $\mathbf{z} - W(\mathbf{x})\mathbf{x}$, enforcing local linearity without incurring the cost of explicit Jacobian construction.
All other components of ILLUME remain unchanged in \illora, including similarity-preserving losses, orthogonality and decorrelation regularization, surrogate training, and rule decoding. 

As a result, \illora does not compromise the performance of \ILLUME, enabling scalable, stable, and biologically interpretable explanations for high-dimensional transcriptomic models.

\subsection{End-to-end XAI pipeline} 
We design a multi-stage framework that combines data-driven prediction with post-hoc explainability to characterize the molecular determinants of drug response.
Figure~\ref{fig:pipeline} (\textbf{b}) summarizes the pipeline. 
First, we preprocess continuous drug response data to define a classification task for each drug by discretizing the values into three ordered bins, corresponding to \emph{sensitive}, \emph{intermediate}, and \emph{resistant} cell lines.
Then, we perform an \emph{all-relevant} feature selection~\cite{kursa2010feature} to identify an informative subset of genes for each drug without imposing arbitrary thresholds and train a gradient-boosting predictive model~\cite{ke2017lightgbm} on the selected features. 
Finally, we resort to \illora to interpret the obtained predictions and extract biological insights.

\spara{Stage 0: Target discretization.} \label{sec:pre-processing}
Drug response is commonly measured by the half-maximal inhibitory concentration ($IC_{50}$), i.e., the drug concentration required to inhibit a biological process by 50\%.
While $IC_{50}$ is traditionally modeled as a continuous outcome using regression~\cite{carli2025learning, iorio2016landscape}, we recast drug response prediction as a classification task to obtain more interpretable explanations. 
In particular, classification enables the extraction of simple decision rules (e.g., \textit{``if gene A expression exceeds $c$ and gene B expression exceeds $d$, then the model predicts sensitivity''}), whereas regression explanations are typically tied to specific numerical predictions rather than broad regimes and are therefore less intuitive. 
From a biomedical perspective, this discretization is not restrictive, as the primary goal is often to distinguish sensitive from resistant cell lines rather than to estimate exact response values and it can, in fact, reduce sensitivity to experimental noise in $IC_{50}$ measurements.
Thus, for each drug $d$, we discretize response values into $\labely_i^{(d)} \in \{1,2,3\}$ using tertile-based binning, yielding three ordered classes: \emph{sensitive}, \emph{intermediate}, and \emph{resistant}.
Class boundaries are computed exclusively on the training data and then applied to evaluation data to prevent information leakage.  
By construction, this procedure produces 
approximately balanced classes, promoting stable training and more reliable evaluation across response classes.  
Alternative discretization strategies such as equal-width or clustering-based binning produce highly imbalanced class distributions, resulting in less stable models and degraded explanations. Moreover, compared to alternative quantile-based splits, we observed that tertile-based binning overall provides the best balance between performance and stability (see Appendix~\ref{App:pipeline_sensitivity} for additional details).\\
The procedure yields the drug-specific classification datasets
$
\drugdataset{d} = \{(\feat_i^{(d)}, \labely_i^{(d)})\}_{i=1}^{N_d}, 
$
where $\feat_i^{(d)} \in \mathbb{R}^{p_d}$ denotes the gene expression profile of the $i$-th cancer cell line and $\labely_i^{(d)}$ its sensitivity class.

\spara{Stage 1: Feature selection.}
Transcriptomic datasets contain measurements for approximately $p_d\approx18{,}000$ genes, many of which are irrelevant for drug response. Considering all such genes can degrade predictive performance and reduce the stability and reliability of model explanations. Thus, feature selection is crucial. 
To limit user-injected bias, we adopt a fully data-driven strategy that automatically identifies features carrying predictive signal   
without predefining the number of selected genes or imposing user-defined thresholds.
Specifically, we leverage \BORUTA~\cite{kursa2010feature}, a feature selection algorithm designed to identify all features that are relevant to the prediction task.
Unlike minimal-optimal methods that seek the smallest predictive subset, \BORUTA retains all features that provide useful information for the task, which is a desirable property in transcriptomic settings, where drug response may arise from partially redundant or correlated pathways.

\spara{Stage 2: Black-box predictor training.}
\label{sec:blackbox}
Using the drug-specific gene set selected by \BORUTA\ and the corresponding response labels, we train a black-box classifier aimed at capturing complex nonlinear relationships between gene expression of cancer cell lines and drug sensitivity.
More formally, after preprocessing, the dataset is
$
\tilde{\mathcal{D}}^{(d)} = \{(\tilde{\feat}_i^{(d)}, \labely_i^{(d)})\}_{i=1}^{N_d}, 
$ where $\tilde{\feat}_i^{(d)} \in \mathbb{R}^{m_d}$ has a reduced set of features ($m_d<p_d$) and $\labely_i^{(d)} \in \{1,2,3\}$ denotes the class label. 
Given these inputs, for each drug $d$,  we learn a three-class predictive model $\model{d}$ from $\tilde{\mathcal{D}}^{(d)}$. %
For each class $c$, the model learns scoring functions
\(
\model{d}_c : \mathbb{R}^{m_d} \rightarrow \mathbb{R},
\)
which assign a real-valued score to each class given an input
$\tilde{\mathbf{x}}$.
At inference time, predictions for $\tilde{\mathbf{x}}$ are obtained by selecting the class with the
highest score:
$
\hat{y}^{(d)}(\tilde{\mathbf{x}}) =
\arg\max_{c \in \{1,2,3\}} \model{d}_c(\tilde{\mathbf{x}}).
$\\
We instantiate $\model{d}$ as
a gradient-boosting ensemble of decision trees~\cite{ke2017lightgbm}, selected for its ability to capture nonlinear relationships and feature interactions, but also for practical considerations. Tree ensembles allow the use of TreeSHAP~\cite{lundberg2020local}, the only SHAP implementation that scales to feature spaces with thousands of genes,  allowing a comparison of our approach with a widely adopted XAI baseline. 

\spara{Stage 3: \illora meta-explainer training.}
\label{sec:illora}
This stage aims to characterize the predictive behavior of the black-box model through structured, biologically interpretable explanations.
Given a trained model $\model{d}$ learned from $\tilde{\mathcal{D}}^{(d)}$, \illora learns a drug-specific explainer
$
\explainer^{(d)} : \big(\model{d}, \tilde{\feat} \big) \;\mapsto\; \explanation^{(d)}(\tilde{\feat}),
$
where $\tilde{\feat} \in \mathbb{R}^{m_d}$ denotes the (filtered) transcriptomic profile and $\explanation^{(d)}(\tilde{\feat})$ is a local explanatory object that describes the decision of the model.
Once trained, \illora 
can provide
multiple complementary forms of explanations, including feature attributions and decision rules.

\medskip
The three stages of the pipeline deliver locally faithful surrogate models that approximate the behavior of the black-box predictor in the neighborhood of individual samples. These surrogates are then analyzed to derive actionable insights and reveal the biological mechanisms driving the model's predictions.

\subsection{Univariate explanations: gene-level feature attribution} \label{subsec:univariate expl}

Gene-specific importance scores are derived from the parameters of the surrogate model learned in \illora latent space, which provides a locally interpretable approximation of the black-box predictor. In this work, the surrogate is instantiated as a linear logistic regression classifier or as a decision tree. 
In the case of the logistic regression classifier, 
the relevance $\psi_i$ of gene $g_i$ can be quantified directly by combining model coefficients with the local linear mappings
$
    \psi_{i}(\mathbf{x}) = \sum_{l=1}^v \beta_l W_{li}(\mathbf{x}),
$
where the coefficients $\beta_l$ capture feature attribution in the latent space surrogate and the weights $W_{li}(\mathbf{x})$ project it back onto the original gene features. 
This formulation yields a transparent estimate of the influence of each gene on the predicted drug response, allowing us to identify genes that systematically promote sensitivity or resistance across samples.

\spara{Evaluation.} 
To assess biological grounding and technical quality of the feature attributions, we use the following standard metrics:
\begin{itemize}[leftmargin=*]

\item \textit{Putative target recovery.}
We evaluate the capability to recover the putative target of the drugs in the 
induced feature ranking through the binary normalized discounted cumulative gain~\cite{jarvelin2002cumulated}:
$$
\small
\mathrm{NDCG}@k
= \sum_{j=1}^{k} \frac{\mathrm{rel}_{g_j}}{\log_2(j+1)} \left[ \sum_{j=1}^{\min(k,\rho)} \frac{1}{\log_2(j+1)} \right]^{-1},
$$
where 
$g_j$ is the gene ranked at position $j$, $\rho$ is the number of putative genes and $\mathrm{rel}_{g_j} \in \{0,1\}$.

\item \textit{Gene set enrichment.} Gene enrichment analysis~\cite{subramanian2005gene} supports biological interpretation by assessing whether selected gene sets show overrepresentation of known pathways. 
In particular, the unweighted gene enrichment score of pathway $\mathcal{S}$ is defined by
$$ \small 
\mathrm{ES}_{\mathcal{S}} 
=
\max_{1 \le \ell \le m}
\left|
\sum_{j=1}^{\ell}
\left(
\frac{\mathbb{1}[g_j \in \mathcal{S}]}{|\mathcal{S}|}
-
\frac{\mathbb{1}[g_j \notin \mathcal{S}]}{m-|\mathcal{S}|}
\right)
\right|,
$$ where  $m$ is the number of ranked genes. 

\item \textit{Explanation robustness.} We evaluate the robustness of feature attribution as the minimum cosine similarity between importance vectors within a local neighborhood of a given instance $\mathbf{x}_i$~\cite{piaggesi2025explanations}: 
$$ {\small \mathrm{Robu}_k(e(\mathbf{x}_i)) = \mathrm{min}_{j \in \mathcal{N}_k^=(i)} \cos \big(e(\mathbf{x}_i), e(\mathbf{x}_j)\big)},$$ where $\mathcal{N}^=_K(i)$ denotes the set of $k$ nearest neighbors of $\mathbf{x}_i$ with the same predicted label. 
\end{itemize}

\subsection{Beyond univariate explanations: from rules to gene-gene interactions}
\label{subsec:beyond univariate}
To derive higher-order, human-interpretable explanations, we extract for each cell line factual decision rules using tree-based surrogate models within \illora. 
Each rule corresponds to a root-to-leaf path in a decision tree trained to locally approximate the black-box behavior in the latent space~\cite{piaggesi2025explanations}, and  takes the form of a conjunction of threshold conditions on gene expression values, $x_j  \in [x^{(low)}_j, x^{(up)}_j],$ where bounds are defined over the domain of $x_j$, extended with $\pm \infty$~\cite{guidotti2024stable}.
For example, a rule may be $x_i > 0.5 \land x_j < 1 \Rightarrow \textrm{Sensitive}$.
By combining multiple constraints, this representation naturally captures joint expression patterns and potential interaction effects driving the surrogate's predictions.

To systematically identify gene interactions within each sensitivity class, we analyze gene co-occurrence across the extracted rule set using the \emph{lift} measure~\cite{tuffery2011data}. 
Let $P(g_i)$ and $P(g_j)$ denote  the marginal frequencies of genes $g_i$ and $g_j$ across rules, and $P(g_i,g_j)$  their joint frequency of co-occurrence; lift is then defined as:
$
\mathrm{lift}(g_i,g_j) = \frac{P(g_i,g_j)}{P(g_i)\,P(g_j)}.
$
Values exceeding $1$ indicate that the two genes co-occur more often than expected under independence, suggesting that their joint presence carries complementary predictive information not reducible to either gene alone.
Ranking gene pairs by lift enables efficient identification of the most relevant interactions among the $\mathcal{O}(m^2)$ candidates.
Importantly, these interactions are model-derived explanatory signals rather than ground-truth biological associations. Our framework deliberately refrains from imposing a priori biological assumptions precisely to prevent biasing the discovery process. Biological validity is instead assessed a posteriori: extracted interactions are treated as testable hypotheses to be evaluated against independent experimental evidence.

To  characterize the global structure induced by the recovered pairwise interactions among genes, we construct the \textit{gene-gene interaction} graph \(\mathcal{G} =(V,E,w)\), where nodes correspond to genes, and each edge \((g_i,g_j)\in E\) is assigned a weight $w(g_i, g_j)$ equal to the lift of the corresponding gene pair.
This graph-based representation naturally supports a range of informative analyses. %

\spara{Evaluation.} 
We use multiple approaches to evaluate the technical quality and biological grounding of gene–gene explanations.

\begin{itemize}[leftmargin=*]
    \item \textit{Pathway overlap} (PO).
    To test whether the pairs of genes with highest lift capture established biological mechanisms, we quantify 
    their functional proximity through the overlap of the associated pathways.
    Formally, let $\Pi$ denote a collection of annotated pathways, each represented as a set of genes. For a gene $g$, define then the aggregated pathway gene set $\Gamma(g) = \bigl\{ h \mid \exists \, \pi \in \Pi \text{ such that } g\in \pi, h \in \pi \bigr\}$, i.e., the set of all genes that belong to at least one pathway containing $g$.
    For a pair of genes $(g_i, g_j)$, we define the pathway overlap as the Jaccard similarity between their aggregated pathway gene sets
    $\mathrm{PO}(g_i, g_j) = \frac{ \lvert \Gamma(g_i) \cap \Gamma(g_j) \rvert }{  \lvert \Gamma(g_i) \cup  \Gamma(g_j) \rvert }$. 
  
\item \textit{Putative target centrality.} 
In the gene-gene interaction graph,  
we evaluate whether 
drug putative targets hold a central position in the graph using different node centrality measures~\cite{freeman1978centrality}.

\item \textit{Pathway conductance.}
In the gene-gene interaction graph, we also assess whether broader biologically connected sets of genes form cohesive sub\-structures within this graph by measuring the weighted conductance. Formally, given the gene set \( \mathcal{S} \subseteq V\) and its complement $\bar{\mathcal{S}} = V \setminus \mathcal{S}$, we evaluate the conductance score
$
\phi(\mathcal{S} )=\frac{w(\partial \mathcal{S} )}{\min\big(\mathrm{vol}(\mathcal{S} ),\,\mathrm{vol}(\bar{ \mathcal{S}} )\big)},
$
where 
\(
w(\partial \mathcal{S} )=\sum_{g_i \in \mathcal{S} , g_j \notin \mathcal{S} } w(g_i,g_j), 
\) 
and $
\mathrm{vol}(\mathcal{S} )=\sum_{g_i\in \mathcal{S},g_j \in V} w(g_i,g_j)$. 
Low conductance indicates that the gene set is more tightly connected internally than externally, reflecting a cohesive substructure in the interaction graph.
\end{itemize}

\section{Results}
\label{sec:results}
In the following, we focus on the evaluation and interpretation of explanations for the sensitive and resistant classes, whose biological signatures are more clearly defined than those of the intermediate class. Moreover, we analyze the scalability of the proposed \illora against the baseline \ILLUME.
To ensure biological validity, we restrict the analysis to drugs whose annotated putative targets are recovered by the feature-selection process, yielding $20$ datasets with transcriptomic profiles for around $500$ cell lines each. 
Descriptions of data sources are provided in Appendix~\ref{App:Dataset}, while preprocessing procedures used for model training and validation are detailed in Appendix~\ref{App:Data_preprocessing}. In addition, 
Appendix~\ref{App:pipeline_sensitivity} reports an ablation study of the proposed XAI pipeline, assessing the effects of substituting its main components with alternative methods. 
Furthermore, Appendix~\ref{app: Biological relevance of gene-gene ranking}-\ref{App:polarity graph} report supplemental experiments on the evaluation of bivariate explanations.
Finally, Appendix~\ref{App:Additional datasets} demonstrates the applicability of our approach across additional drug datasets.

\subsection{Single-gene attribution analysis}

\begin{figure}[t]
\centering
    \begin{tabular}{c|c}
        \includegraphics[width=0.45\linewidth]{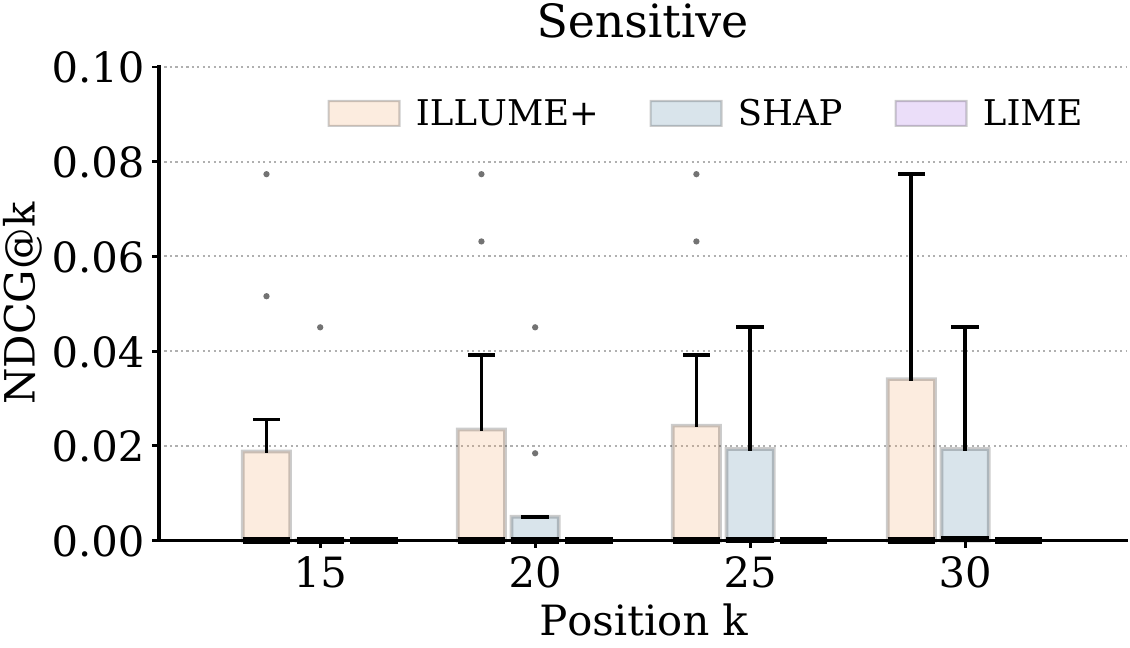}  &  
         \includegraphics[width=0.45\linewidth]{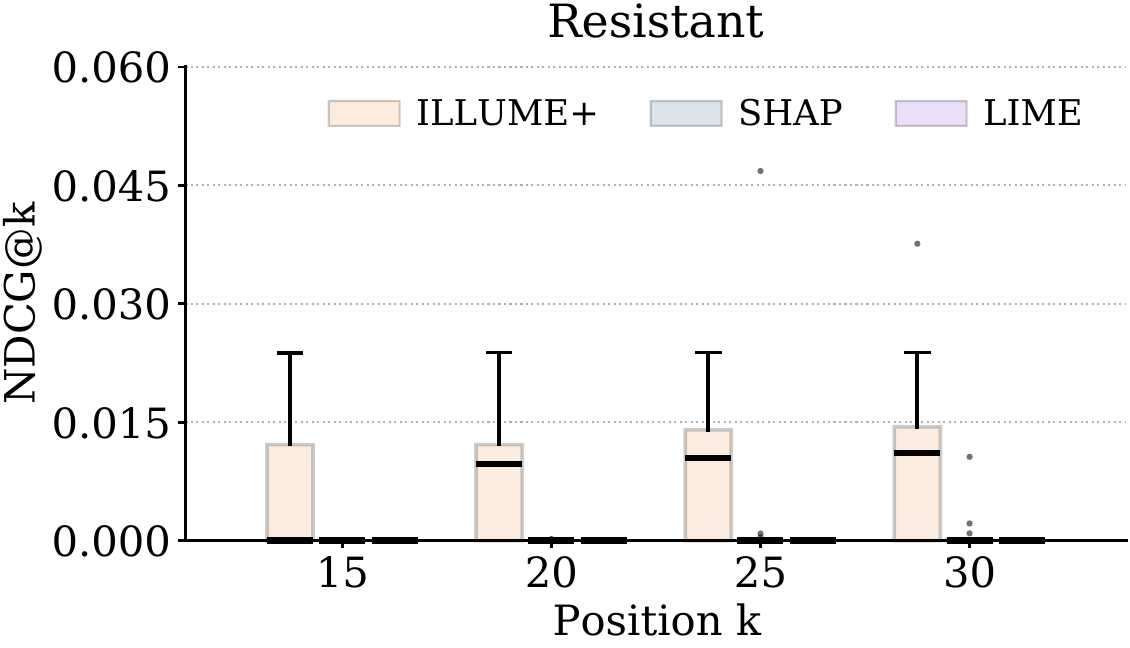} \\
    \end{tabular}
    \caption{Putative gene recovery (measured via NDCG@k) across different values of $k$  for \sensitive~and \resistant~classes.} 
    \label{fig:ncgdatk}
\end{figure}

\begin{figure}[t]
    \centering
    \begin{tabular}{c@{\hspace{0.4cm}}|@{\hspace{0.4cm}}c}
    \includegraphics[width=0.35\linewidth]{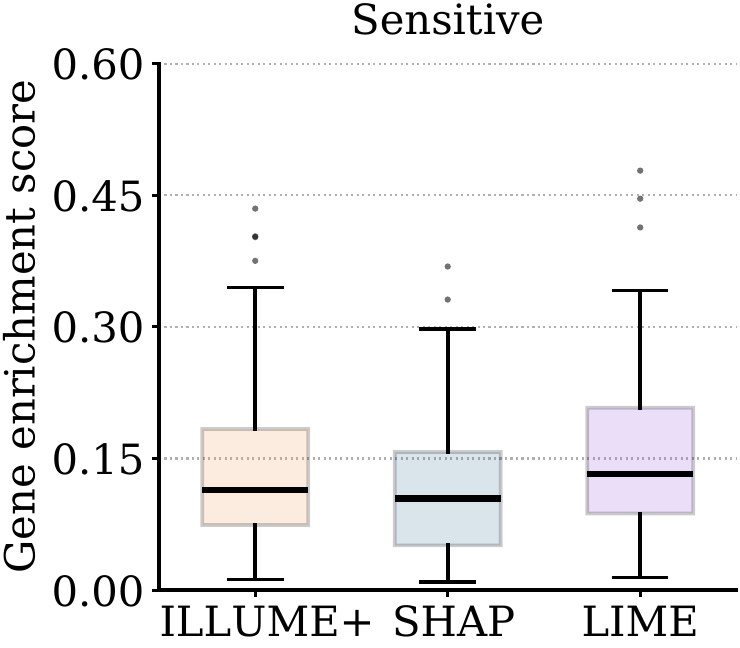} 
    \hspace{0.25cm}
    &
    \hspace{0.25cm}
    \includegraphics[width=0.35\linewidth]{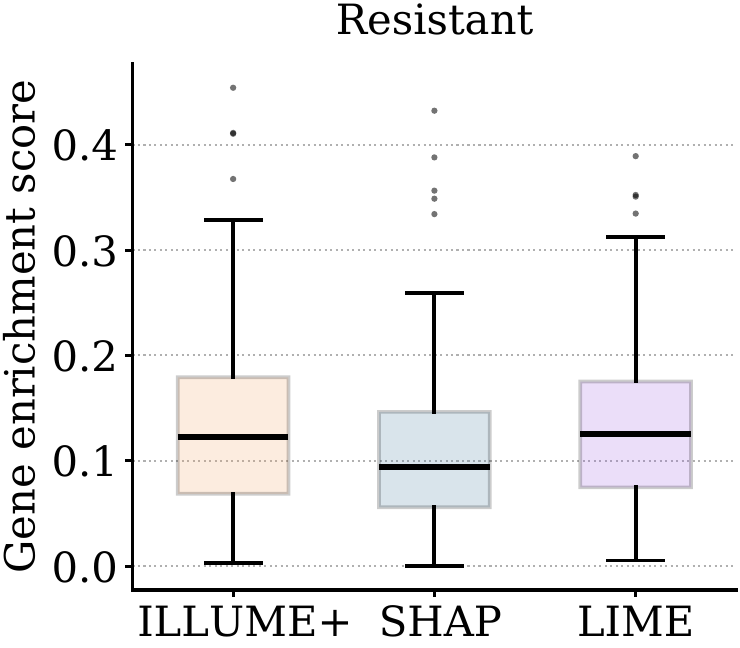} \\
    \end{tabular}
    \caption{Gene enrichment scores across pathways and drugs for sensitive and resistant classes.
    }
    \label{fig:enrichment_score}
\end{figure}

\begin{figure}[t]
\centering
    \begin{tabular}{c|c}
        \includegraphics[width=0.45\linewidth]{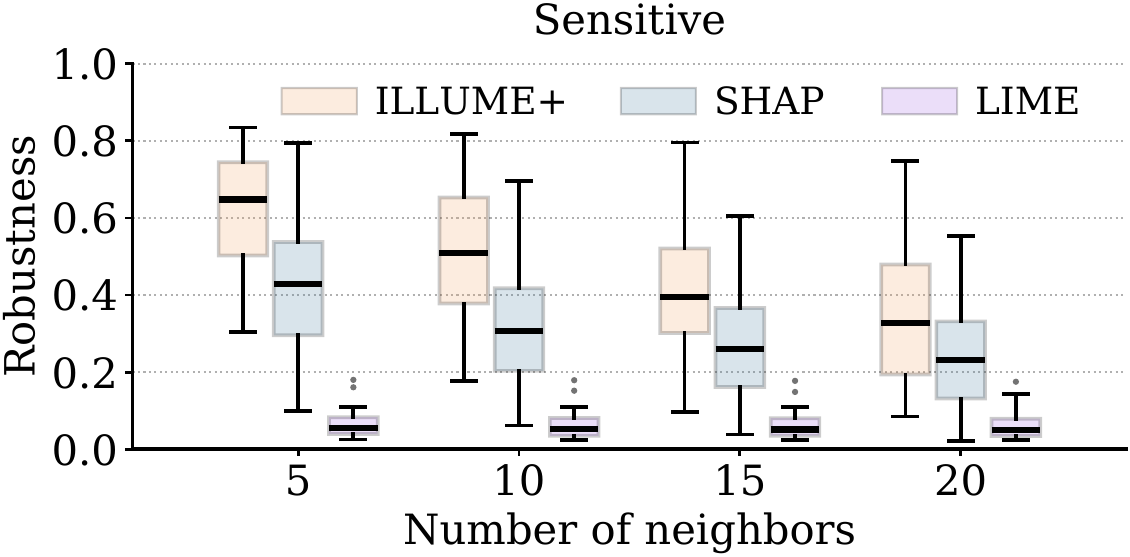}  &  
         \includegraphics[width=0.45\linewidth]{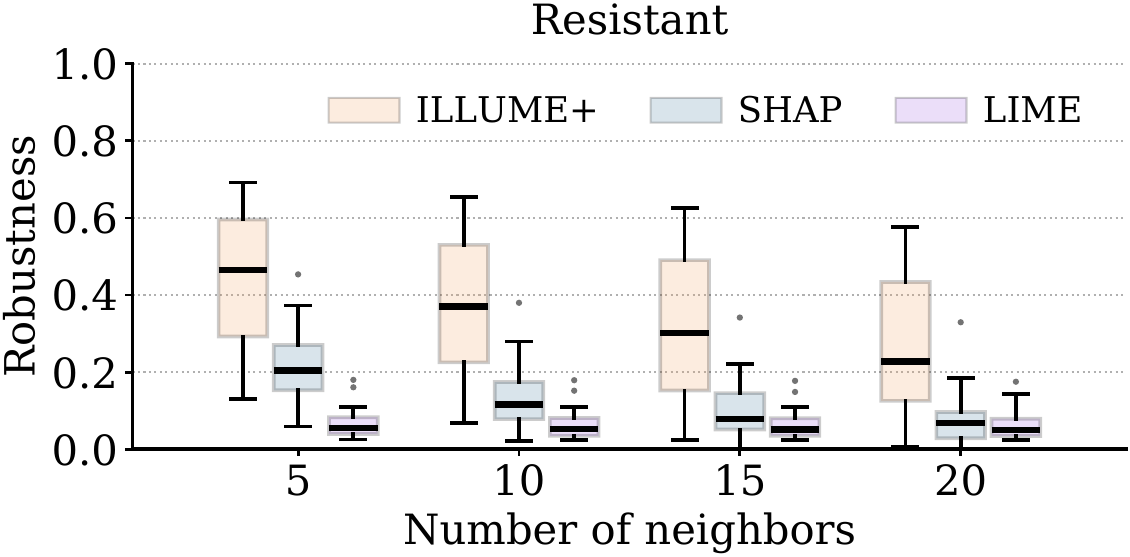} \\
    \end{tabular}
    \caption{Robustness (measured via cosine similarity) across different values of $k$ for \sensitive \ and \resistant  \ classes.}
\label{fig:robustness_by_k_sensitive}
\end{figure}

\begin{figure*}[t]
\includegraphics[trim={0 0 0 3mm},clip, width=0.925\linewidth]{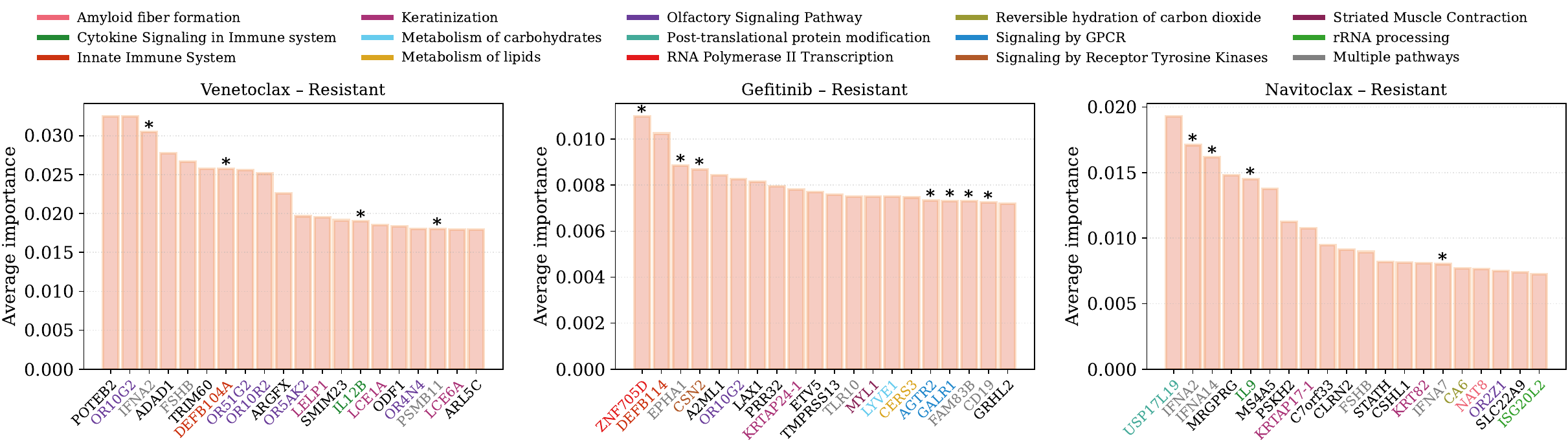}
\caption{
Pathway-colored top-20 gene importance profiles for three example drugs for the resistant class. 
Genes associated with known mechanisms of action are marked with a star sign.}
\label{fig:pathway_colored_top20_5drugs}
\end{figure*}

As discussed in Section~\ref{sec:introduction}, prior XAI studies for drug response prediction mainly rely on univariate attribution methods such as SHAP. 
Therefore, we start by comparing the biological grounding as well as internal robustness of univariate feature attributions from \illora with those from SHAP, and include as additional baseline LIME~\cite{ribeiro2016should}, an alternative popular post-hoc explainer.
\Cref{fig:ncgdatk} reports NDCG@$k$ values across drugs as a function of $k$, indicating that \illora more effectively ranks putative targets among the top genes.
We further evaluate biological relevance through enrichment of MoA-related pathways derived from Reactome~\cite{milacic2024reactome}, i.e., all pathways including the drug or its putative targets. 
As shown in \Cref{fig:enrichment_score}, which aggregates scores across drugs and pathways, \illora prioritizes MoA-related genes more effectively than SHAP (paired Wilcoxon signed-rank test; $p$-value$=0.046$) and yields results comparable to LIME ($p$-value$>0.1$).
To illustrate the models' internal consistency, we report the values of the cosine-similarity-based robustness metric as the number of neighbors varies in \Cref{fig:robustness_by_k_sensitive}, demonstrating that \illora yields more robust explanations than both SHAP and LIME across the entire range of neighborhood sizes.
Overall, \illora offers the best trade-off between biological relevance and robustness: LIME attains similar enrichment but lower robustness and target recovery, while SHAP performs significantly worse across all metrics.
To provide a more concrete illustration, in \Cref{fig:pathway_colored_top20_5drugs} we illustrate the top-20 ranked genes for three representative drugs, with each gene annotated by its associated pathway.

\subsection{Rules and gene-gene interaction analysis} 
In this section, we report results of our investigation of the interactions between genes based on the rules extracted by \illora. 

\begin{figure}[t]
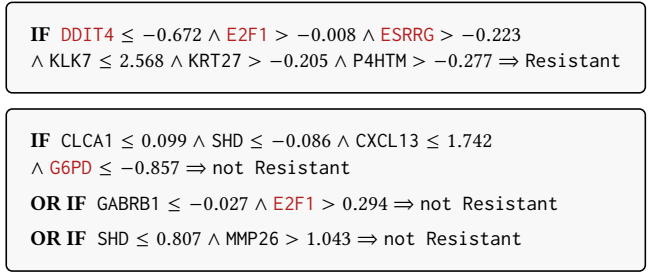

	\centering
	
	\begin{tcolorbox}[rulebox]
		\small\raggedright
		\textbf{IF}\;
		\textcolor{pathcol}{\texttt{DDIT4}} \(\leq -0.672\)
		\(\wedge\)
		\textcolor{pathcol}{\texttt{E2F1}} \(> -0.008\)
		\(\wedge\)
		\textcolor{pathcol}{\texttt{ESRRG}} \(> -0.223\)\\
		\(\wedge\)
		\texttt{KLK7} \(\leq 2.568\)
		\(\wedge\)
		\texttt{KRT27} \(> -0.205\)
		\(\wedge\)
		\texttt{P4HTM} \(> -0.277\)
		\(\Rightarrow\)
		\texttt{Resistant}
	\end{tcolorbox}
	
	\begin{tcolorbox}[rulebox]
		\small\raggedright
		\textbf{IF}\;
		\texttt{CLCA1} \(\leq 0.099\)
		\(\wedge\)
		\texttt{SHD} \(\leq -0.086\)
		\(\wedge\)
		\texttt{CXCL13} \(\leq 1.742\)\\
		\(\wedge\)
		\textcolor{pathcol}{\texttt{G6PD}} \(\leq -0.857\)
		\(\Rightarrow\)
		\texttt{not Resistant}
		
		\medskip
		\textbf{OR IF}\;
		\texttt{GABRB1} \(\leq -0.027\)
		\(\wedge\)
		\textcolor{pathcol}{\texttt{E2F1}} \(> 0.294\)
		\(\Rightarrow\)
		\texttt{not Resistant}
		
		\medskip
		\textbf{OR IF}\;
		\texttt{SHD} \(\leq 0.807\)
		\(\wedge\)
		\texttt{MMP26} \(> 1.043\)
		\(\Rightarrow\)
		\texttt{not Resistant}
	\end{tcolorbox}
	
	\caption{
		Examples of factual (top) and counterfactual (bottom) rules for \drug{Alisertib}. Genes in red belong to the \emph{RNA Polymerase II Transcription} pathway.
	}
	\label{fig:pathway_rule_example}
\end{figure}

\spara{Visualization of decision rules.}
Interpretable decision rules provide an intuitive description of how transcriptomic patterns drive the predicted cancer drug response, combining genes from established biological pathways with additional data-driven associations that extend beyond existing knowledge. 
Importantly, for extracted rules to be actionable and cognitively interpretable, they must remain sufficiently concise. 
To this end, \illora explicitly enforces sparsity during training by constraining each latent feature to be a linear combination of at most $2$ input attributes (gene expressions). This design naturally yields concise and interpretable rules.
Figure~\ref{fig:pathway_rule_example} illustrates a representative rule containing broadly prognostic genes (e.g., \drug{DDIT4}~\cite{pinto2017silico}, \drug{E2F1}~\cite{li2023prognostic}, \drug{KLK7}~\cite{kind2024klk7}), tissue-specific markers such as \drug{P4HTM}~\cite{didonna2023p4htm}, and genes functionally linked to MoA-related pathways (e.g., \drug{DDIT4}, \drug{E2F1}, \drug{ESRRG}). The distribution of rule lengths and more example rules are reported in the Appendix~\ref{app:decision_rules}.
In addition, \illora extracts concise counterfactual rules determining which changes in gene expressions would modify the predicted class. The recurrence of \drug{E2F1} across both factual and counterfactual rules suggests a context-dependent role, depending on the expression state of accompanying genes.
To the best of our knowledge, the combined association with drug response of multiple retrieved genes has not been previously reported, pointing to testable biological hypotheses. Inspection of other extracted rules revealed the recurrent involvement of genes with documented roles in cancer prognosis.
For instance, for the drug \drug{Erlotinib}, a receptor tyrosine kinase inhibitor (RTKI) targeting the Epidermal Growth Factor Receptor (EGFR), we found the following rule:
\begin{align*}
&\pg{CNTNAP1}\!>\!-0.891~\wedge~\pg{EVL}\!\leq\!-1.395~\wedge~\gene{FAM214B}\!\leq\!0.212~\wedge 
 \\
 &\gene{IL13RA1}\!\leq\!0.976~\wedge~\gene{IRF6}\!>\!0.390~\wedge~\pg{PSEN2}\!>\!0.133
 \;\Rightarrow\; \text{\textsc{Resistant}} 
\end{align*}
High levels of PSEN2, a core component of the $\gamma$-secretase complex required for Notch activation, is consistent with established Notch-mediated \drug{Erlotinib} resistance~\cite{mur2020notch}.
Similarly, reduced EVL levels have been associated to colorectal cancer ~\cite{yu2023elevated} and increased metastasis in breast cancer~\cite{padilla2018actin}. Other contributions are more nuanced: IRF6~\cite{muralidharan2024breast, botti2011developmental} and IL13RA1 ~\cite{bednarz2020interleukins, shi2022loss} have tumor-dependent roles.
Other genes have not been thoroughly investigated in cancer, but could be potential biomarkers: among them, CTNAP1 has been suggested as a clear cell renal cell carcinoma marker in an independent bioinformatic analysis~\cite{li2022m2}.

While counterfactual rules provide complementary insights into transcriptomic patterns associated with drug sensitivity, 
we leave their systematic characterization to a future work and focus, in this paper, on the analysis of factual rules.

\begin{figure*}
\centering
\includegraphics[trim={0 45mm 0 0},clip,width=0.95\linewidth]{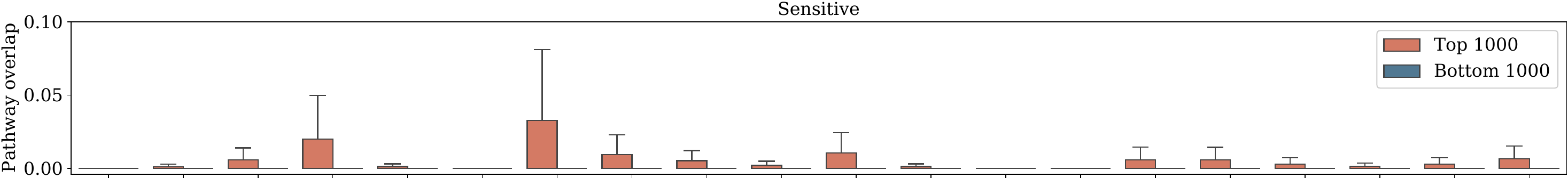} \\
\includegraphics[trim={0 0 0 0},clip,width=0.95\linewidth]{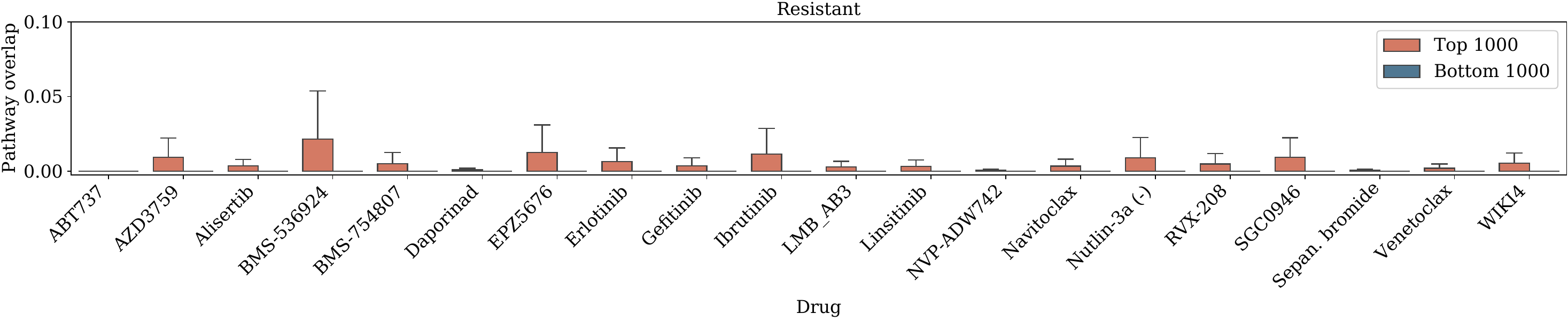}
\caption{Distribution of pathway overlap across drugs for the top (left box plot) and bottom (right box plot) 1,000 ranked pairs.}
\label{fig:interaction_pathway_overlap_by_drug}
\end{figure*}

\begin{figure}
    \centering
    \includegraphics[trim={0 6mm 0 0}, clip, width=0.8\linewidth]{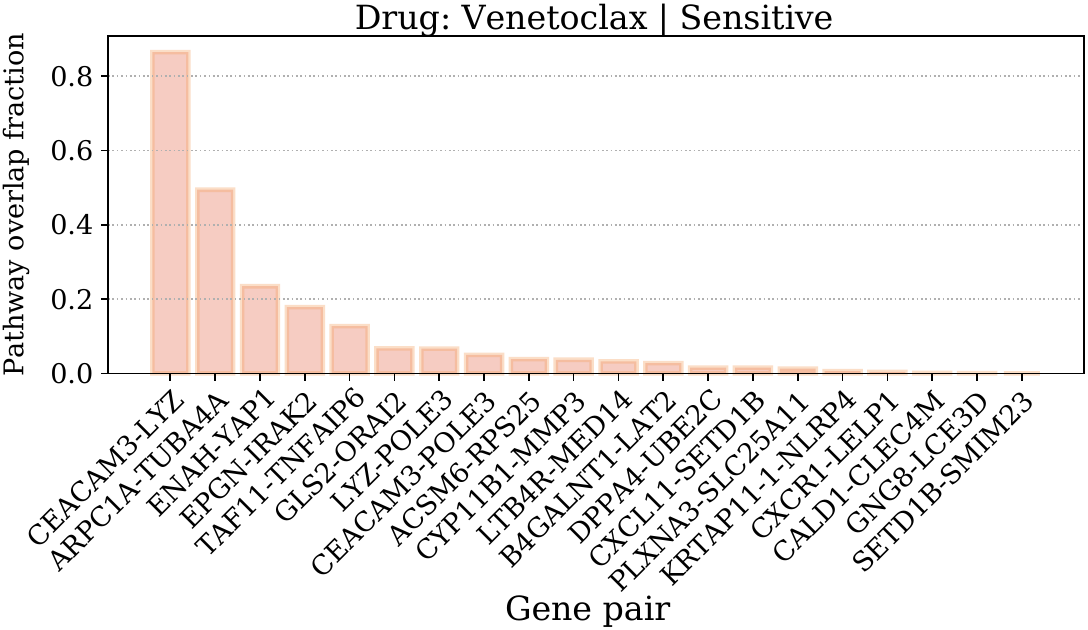} \\
    \includegraphics[width=0.8\linewidth]{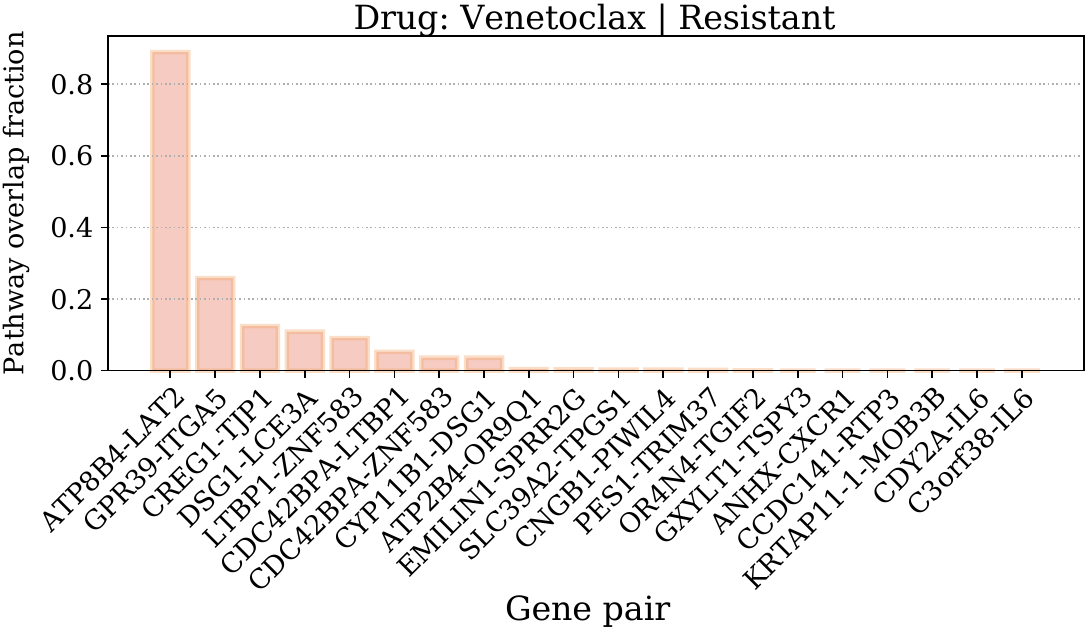} 
    \caption{Pathway overlap for the top-20 gene pairs in \drug{Venetoclax} \sensitive (top) and \resistant (bottom) classes. 
  }  \label{fig:pathway_colored_top20_5drugs_pairwise} 
\end{figure}

\spara{Gene-gene interaction ranking analysis.}
To assess the biological relevance of the interaction ranking, we computed the pathway overlap (PO) of top-ranked gene pairs and compared it against that of the bottom 1,000 pairs. Figure~\ref{fig:interaction_pathway_overlap_by_drug} reports the distribution of PO values across drugs, revealing a clear separation between highly ranked and low-ranked interactions.
Because PO values are broadly dispersed and include several extreme observations, they are omitted from the figure for readability. 
Our approach yields significantly higher pathway overlap for top-ranked than for bottom-ranked gene pairs for all drugs (Mann--Whitney U test, $p$-value$<0.05$). In contrast, the same analysis reaches significance for only 40\% of drugs when rankings are obtained with 
SHAP's interaction values~\cite{lundberg2020local}, suggesting weaker biological coherence of its pairwise rankings (see Appendix~\ref{app: Biological relevance of gene-gene ranking} for additional details).
Figure~\ref{fig:pathway_colored_top20_5drugs_pairwise} illustrates, for an exemplifying drug, 
the PO scores of the 20 pairs with largest pathway overlap, among the 50 pairs with largest lift. 
Here, we see that gene pairs with large PO and large importance value point to different cancer-related pathways in sensitive and resistant cells. In particular, interactions in \drug{Venetoclax}-sensitive cells involve genes in inflammatory response and neutrophil chemotaxis, pointing to exposure of cancer cells to the immune system (which increases susceptibility).  On the other hand, important pairs in resistant cells are involved in angiogenesis and in cell adhesion, pathways that support tumor viability and invasion.
In Appendix~\ref{app: Biological relevance of gene-gene ranking}, we complement the main analysis with further evidence considering additional drugs as well as alternative measures of relevance for pairwise rankings.

Overall, top-ranked gene pairs identified by our method recover well-established cancer-related pathways, despite the absence of any explicit knowledge-grounded supervision during training, demonstrating that biologically meaningful signals emerge as a by-product of the model's predictive reasoning.

\begin{figure}[t]
    \includegraphics[trim={0 35mm 0 0}, clip, width=\linewidth]{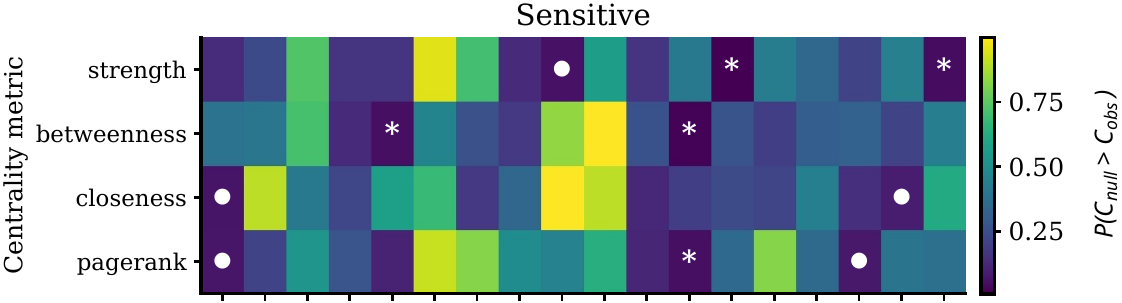}  \includegraphics[trim={0 0 0 0}, clip, width=\linewidth]{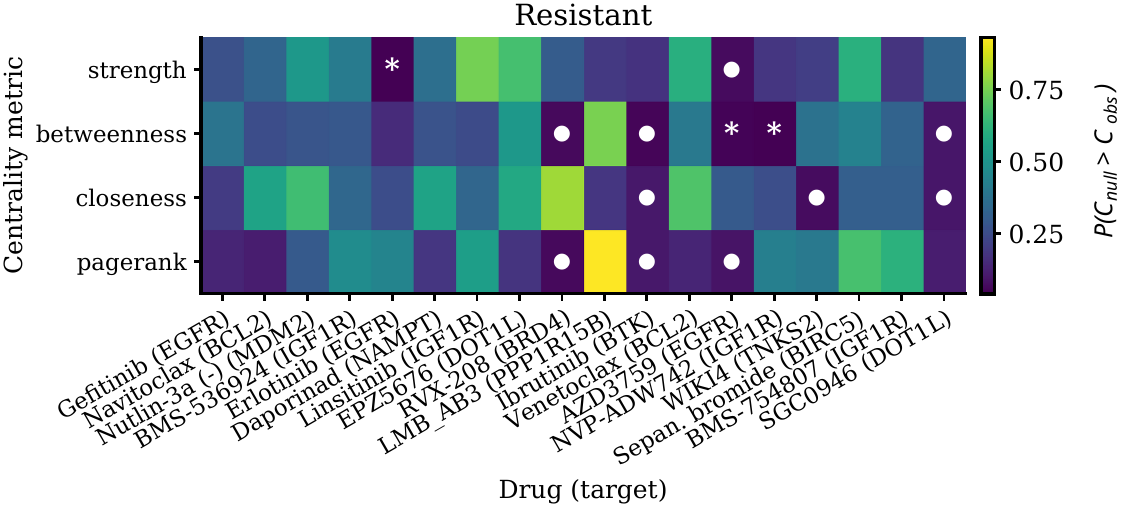} \\
\caption{Empirical p-values for centrality measures of the putative target.  
A dot marks results significant at 
$10\%$
level, while an asterisk marks results significant at the 
$5\%$
level.
}
\label{fig:centrality}
\end{figure}

\spara{Gene-gene interaction graph analysis.} 
Given the gene–gene interaction graph  \(\mathcal{G}\), we investigate the extent to which \emph{putative drug targets} occupy systematically central positions within the constructed network. 
Formally, we test whether target genes exhibit higher centrality than expected under an appropriate null model.
Therefore we compute complementary weighted centrality measures for all nodes: (i)  \emph{strength} \(s(u)=\sum_{v}w(u,v)\), (ii) \emph{PageRank}, and (iii) path-based metrics (\emph{closeness} and \emph{betweenness}), where stronger interactions correspond to shorter paths~\cite{freeman1978centrality}. 
Statistical significance is assessed via one-sided empirical tests tailored to the hypothesis that target genes may be more central than expected under a null model.
Because centrality measures are often correlated with node strength, we adopt a strength-matched null model: genes are first grouped into quantile bins based on their strength, and $1000$ null genes are then generated by uniformly sampling from the same bin as the target gene. 
We apply this strength-matched null to closeness, betweenness and PageRank centrality.
For strength itself,
we instead sample nodes uniformly at random.  
In all cases, the empirical right-tailed $p$-value is computed as the proportion of null genes whose centrality is greater than or equal to that of the target gene.
Results in Figure~\ref{fig:centrality} indicate that various putative targets occupy prominent network positions according to their role. For instance, targets involved in chromatin remodeling and global transcriptional regulation (e.g., \drug{BRD4}, \drug{DOT1L}) display high strength in sensitive lines, consistent with hub-like behavior, whereas their reduced strength in resistant lines suggests extensive chromatin deregulation, a hallmark of aggressive cancers. 
Targets in signalling pathways such as \drug{BCL2} and \drug{EGFR} show elevated betweenness for drugs with a single target (\drug{Venetoclax}, \drug{Erlotinib}), consistent with them being a key information transfer node; this applies to sensitive but not resistant lines, which may have bypassed the signalling cascade.

\begin{figure}[t]
    \centering
    \begin{tabular}{c|c}
    \includegraphics[width=0.45\linewidth]{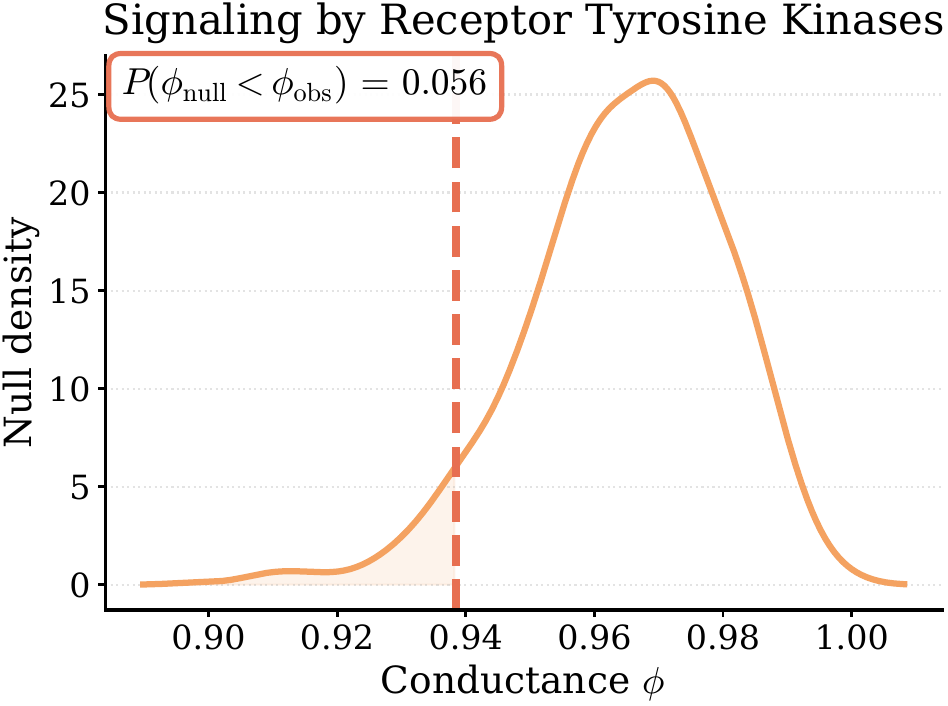}   &
    \includegraphics[width=0.45\linewidth]{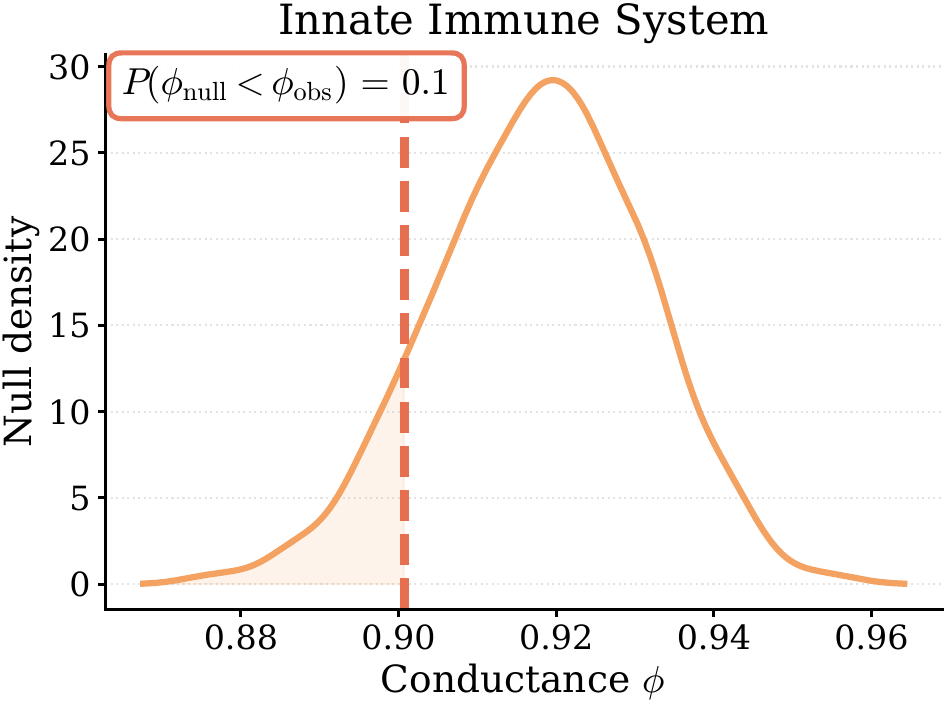}  \\
    \end{tabular}
    \caption{
    Null distribution of conductance and empirical conductance of the pathway specified above for the \sensitive class to the drug  \drug{Navitoclax} (left) and the \resistant class to \drug{Erlotinib}  (right). 
    }
    \label{fig:example_graph}
\end{figure}

We next examine whether biologically related gene sets form cohesive substructures within the interaction graph. To this end, we evaluate the conductance of genes from MoA-related pathways, which quantifies how much genes involved in the drug mechanism of action preferentially interact with one another rather than with the rest of the graph.
Because conductance depends on node degree, we assess significance using a degree-adjusted null model.
In practice, genes are binned by strength in $30$ bins, and $1000$ degree-matched gene sets are generated by sampling replacements from the same strength bins, preserving pathway size and approximate total strength.
For each pathway  with at least five genes present in \(\mathcal{G}\), 
we compute a one-sided empirical
$p$-value by comparing the observed pathway score to a null distribution obtained from degree-matched random gene sets.
The 
$p$-value is defined as the proportion of random sets whose score is less than or equal to that observed for the pathway. 
Across drugs and pathways, for the \sensitive (\resistant) class, approximately \(9.5\%\) (\(14\%\))  of pathways exhibit empirical p-value lower than $0.1$ and around  \(8\%\) (\(6.5\%\)) lower than $0.05$.   
Examples are shown in Figure~\ref{fig:example_graph}. 
Strong interactions are not expected to be confined to annotated pathways, since cross-pathway edges may arise from shared upstream regulation or signaling crosstalk. In fact, this framework enables the identification of previously uncharacterized cohesive subgraphs, generating testable biological hypotheses.
In Appendix~\ref{App:polarity graph}, we also provide interaction analyses based on different definitions of graph structure.

\subsection{Scalability analysis}
Considering the \drug{Gefitinib}, \drug{Navitoclax} and \drug{Venetoclax} drugs,  we evaluate the scalability of \ILLoRA and quantify the contribution of its two optimized components: low-rank parameterization and stochastic approximation of the Jacobian regularizer. To assess their impact, Table~\ref{tab:scalability_memory} reports the average peak GPU memory required during meta-explainer training as the number of input features increases.

\begin{table}[t!]
\centering
\caption{Average peak GPU memory usage (GB), during meta-explainer training by number of input features.}
\label{tab:scalability_memory}
\begin{tabular}{lccc}
\toprule
Method & 100 features & 250 features & 500 features \\
\midrule
ILLUME & 0.88 & 2.22 & 4.50 \\
\illora & 0.39 & 0.56 & 1.14 \\
No low-rank & 0.40 & 0.62 & 1.37 \\
No Jacobian approx. & 1.05 & 2.18 & 4.32 \\
\bottomrule
\end{tabular}
\end{table}

The results show that \ILLoRA substantially reduces memory consumption compared with \ILLUME, with the advantage becoming more pronounced as dimensionality increases, and the ablation analysis identifies in the stochastic Jacobian approximation the main driver of these savings.
Regarding runtime, both \ILLUME and \ILLoRA require a one-time training phase for the meta-explainer. Once training is completed, explanations are generated through lightweight local surrogate models and incur negligible computational overhead. Consequently, training-time memory rather than explanation-time latency constitutes the main practical bottleneck.

To verify that the scalability improvements do not compromise explanation quality, we compared the full \ILLoRA model with the two ablated variants on reduced-scale datasets where all configurations remain computationally feasible (i.e., for subsets of $500$ features selected by \BORUTA), evaluating explanation robustness, putative-target recovery (NDCG@30), and surrogate fidelity.
Overall, the three variants exhibit similar performance across all metrics. Relative to the full model, the largest observed decreases are approximately 17\% in robustness (computed using 20 nearest neighbors) and at most 14\% in surrogate fidelity, measured by the average $F_1$ score of decision tree and logistic regression surrogate models across the training, validation, and test sets, while NDCG@30 remains comparable across configurations. These results indicate that the low-rank parameterization and stochastic Jacobian approximation primarily improve scalability, with only limited impact on explanation quality and surrogate fidelity.

\section{Discussion and Conclusions}
\label{sec:discussion}
We have presented an end-to-end data-driven pipeline based on XAI to uncover the decision logic learned by nonlinear models in high-dimensional transcriptomic settings.
While previous studies~\cite{carli2025learning} have shown that SHAP-based explanations can recover known drug-associated genes,
our results indicate that such explanations can be unstable and miss important genes. 
Most importantly, univariate explanations
provide an incomplete view of the underlying mechanisms driving drug response. 
The gene–gene interaction analyses that we carry out show that the learned interaction structure
exhibits pathway-consistent organization and coherent shifts between sensitive and resistant phenotypes.

More broadly, this work demonstrates that principled explainability can shift predictive modeling from a purely performance-driven exercise toward 
an hypothesis-generating tool for AI-assisted scientific discovery.
By enabling robust, scalable, and pairwise gene explanations, our framework supports more faithful auditing of black-box models and provides a foundation for generating biologically-grounded hypotheses. These insights are relevant not only for drug response prediction, but for a wide range of applications at the intersection of machine learning and the life sciences.
Future work may extend our approach to other applications, while also exploring higher-order gene interactions and distilling explanations into minimal, biologically meaningful marker sets for robust patient and sample stratification.

\begin{acks}
This work has been partially supported by the European Community programme under the funding schemes: G.A. 101286379 ``ILLUME-4-Science'', and G.A. 101120763 ``TANGO'' and by the Italian Project Fondo Italiano per la Scienza FIS00001966 ``MIMOSA''.
Authors also acknowledge the European Project ERC-2018-ADG G.A. 834756 ``XAI- Science and technology for the explanation of AI decision making'', and PNRR - M4C2 - Investimento 1.3, Partenariato Esteso (grant No. PE00000013) - ``FAIR - Future Artificial Intelligence Research'' - Spoke 1 ``Human-centered AI'', funded by the European Commission under the Next Generation EU programme.
\end{acks}

\section*{Limitations and Ethical Considerations}
This work is based on publicly available pharmacogenomic data derived from cancer cell lines, and therefore raises no additional concerns regarding data privacy or informed consent beyond those addressed by the original data providers. 
Predictive signals and explanations may reflect dataset-specific biases related to experimental conditions, tissue composition, or feature selection procedures. 
The extracted explanations capture statistical associations learned by the model and require independent experimental validation before being considered indicative of causal biological mechanisms.

\section*{GenAI Disclosure}
A large language model–based chatbot was used  for language editing to improve clarity, grammar, and readability of the manuscript. The tool did not contribute to the study design, data analysis, interpretation of results, or generation of scientific content. All methodological decisions, analyses, and conclusions were developed and verified by the authors.

\bibliographystyle{ACM-Reference-Format}
\balance
\bibliography{references}

\appendix

\setcounter{table}{0}
\renewcommand{\thetable}{A\arabic{table}}

\setcounter{figure}{0}
\renewcommand{\thefigure}{A\arabic{figure}}

\section*{APPENDIX}

\section{Data sources}
\label{App:Dataset}

Transcriptomic data are derived from the \emph{Cancer Cell Line Encyclopedia} (CCLE)~\cite{barretina2012cancer}, a collection of bulk RNA-seq data from 1699 cancer cell lines. We use log2-transformed transcripts per million plus one (TPM+1) RNA-seq data for 18,174 protein-coding genes.
Drug response data are obtained from the \emph{Genomics of Drug Sensitivity in Cancer} (GDSC)~\cite{iorio2016landscape}, a large-scale drug perturbation dataset which provides drug efficacy measurements for $286$ cancer drugs tested across $969$ cancer cell lines. Drug effectiveness in reducing cell viability is evaluated by half-maximal inhibitory concentration ($IC_{50}$) measurements, i.e., the drug concentration required to kill $50\%$ of the cells. 

Here we treat the $IC_{50}$ measurements associated with each drug as defining a separate dataset; accordingly, whenever we refer to a dataset, we mean the collection of $IC_{50}$ responses observed across cell lines for a specific drug and the associated gene expression values. GDSC provides extensive metadata including the tissue of origin for each cancer cell line as defined by Oncotree~\cite{kundra2021oncotree}, and the putative targets and MoA for each drug.  
Throughout this study, we restricted our analysis to drugs whose targets correspond to genes available in the CCLE dataset. This allowed us to later check whether our results were biologically consistent, and resulted in a set of $151$ drugs.

To define pathways associated to drug response, we employ the Reactome pathway database~\cite{milacic2024reactome}, a curated database of human biological pathways and reactions. We define as MoA-related pathways all the pathways including the drug or its putative targets. Throughout our analysis, we use 1st-layer Reactome pathways as in prior work~\cite{carli2025learning}. 

To support our discussion on the gene–gene interaction graph, we also leverage the STRING database, a resource for functional protein networks integrating curated databases, experiments, co-expression, gene context and text mining~\cite{szklarczyk2023string} (see Figure ~\ref{fig:selling_figure} in \Cref{sec:introduction}). STRING includes both known and predicted interactions and aggregates evidence from multiple knowledge sources, offering higher coverage than curated databases and a unified interaction confidence score. We consider only high-confidence interactions (confidence score > 0.7) between genes with high interaction score from \illora. For each gene, we include up to 5 direct interactors from STRING which are not in our high-interaction list, to allow for long-distance connections. 
In the example graph in Figure~\ref{fig:selling_figure}, 4/5 interactors in STRING not retrieved by \illora are genes removed by \BORUTA. Thus, missing interactions reflect prior feature filtering rather than limitations of \illora.

\section{Data pipeline}
\label{App:Data_preprocessing}

\begin{figure}[t]
    \centering
    \includegraphics[trim={0 0 0 10mm},clip,width=0.8\linewidth]{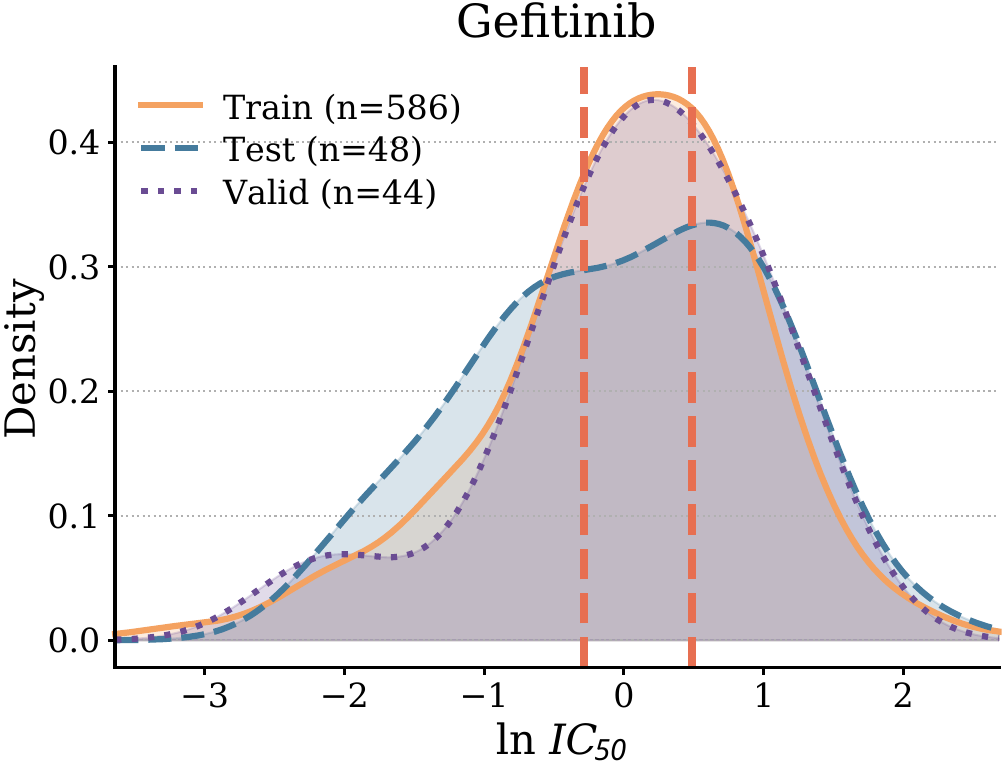} \\
    \includegraphics[trim={8mm 0 15mm 15mm},clip,width=0.8\linewidth]{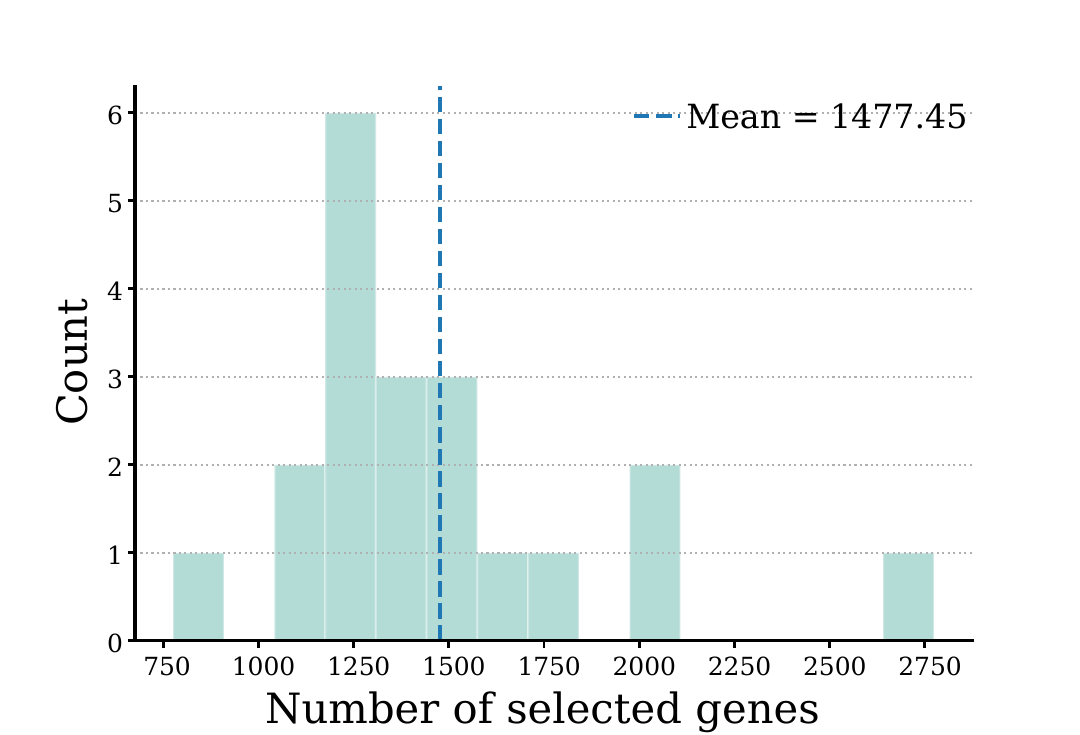} 
    \caption{Label preprocessing and Feature selection. Top: distribution of the target variable ($\ln{IC_{50}}$) across data splits and label boundaries for the three classes as dotted vertical lines in the \textbf{Gefitinib} drug. 
    Bottom: distribution of the number of genes selected by \BORUTA\ across the analyzed drugs.}
    \label{fig:Boruta}
\end{figure}

\spara{Data preprocessing.} 
Each drug dataset is split with approximately 90/10 proportion, with 90\% used for training models, and
the remaining 10\% reserved for testing. To control for tissue-specific confounding effects and ensure fair evaluation, following prior work~\cite{carli2025learning},  data splits are stratified with respect to the tissue of origin.
Drug response, originally measured as continuous ($\ln IC_{50}$) values, is discretized into three classes (\emph{sensitive}, \emph{intermediate}, \emph{resistant}) using tertile-based binning computed over training data.  As an illustration, on the top panel of \Cref{fig:Boruta} we show the distribution of ($\ln{IC_{50}}$) values across data splits for a representative drug, together with the thresholds used to define the three classes.

\spara{Feature selection.} 
\emph{All-relevant} feature selection is performed using the \BORUTA\ algorithm~\cite{kursa2010feature} on each drug training set. 
\BORUTA\footnote{\url{https://github.com/scikit-learn-contrib/boruta_py}} operates by augmenting the original dataset with shadow features, generated by randomly permuting the values of each feature across samples. A Random Forest model~\cite{Breiman01} is then trained on the extended dataset, and feature importance scores are computed. For each iteration, the importance of every real feature is statistically compared to the maximum importance achieved by the shadow features. Features that consistently outperform the shadow features are labeled as \emph{confirmed}, those performing consistently worse are \emph{rejected}, and ambiguous features are classified as \emph{tentative} until a decision can be reached through repeated iterations until early stopping is activated.
This approach provides several advantages in the context of cancer transcriptomics as it reduces dimensionality while limiting the removal of possibly relevant genes and accommodates nonlinear relationships and interactions, which are common in gene expression data.\\
We rely on a class-balanced Random Forest classifier for $3$-class prediction of drug sensitivity levels, and adopt a percentile-based acceptance criterion: a gene is selected if its importance exceeds the $90^{\mathrm{th}}$ percentile of the shadow feature importances. This conservative strategy favors the retention of potentially informative genes while limiting the inclusion of spurious predictors.
Analogously, we retained \emph{confirmed} as well as \emph{tentative} features for downstream modeling and analysis.
We then restrict the analysis to those drugs (20 out of 151) for which the putative targets annotated in the GDSC dataset are fully recovered by this selection procedure. Each selected drug is associated with approximately 500 profiled cell lines. The bottom panel of~\Cref{fig:Boruta} shows the distribution of the selected feature set sizes, which range from roughly 700 to 2,700 genes.

\begin{figure}[h]
    \centering
    \includegraphics[width=0.8\linewidth]{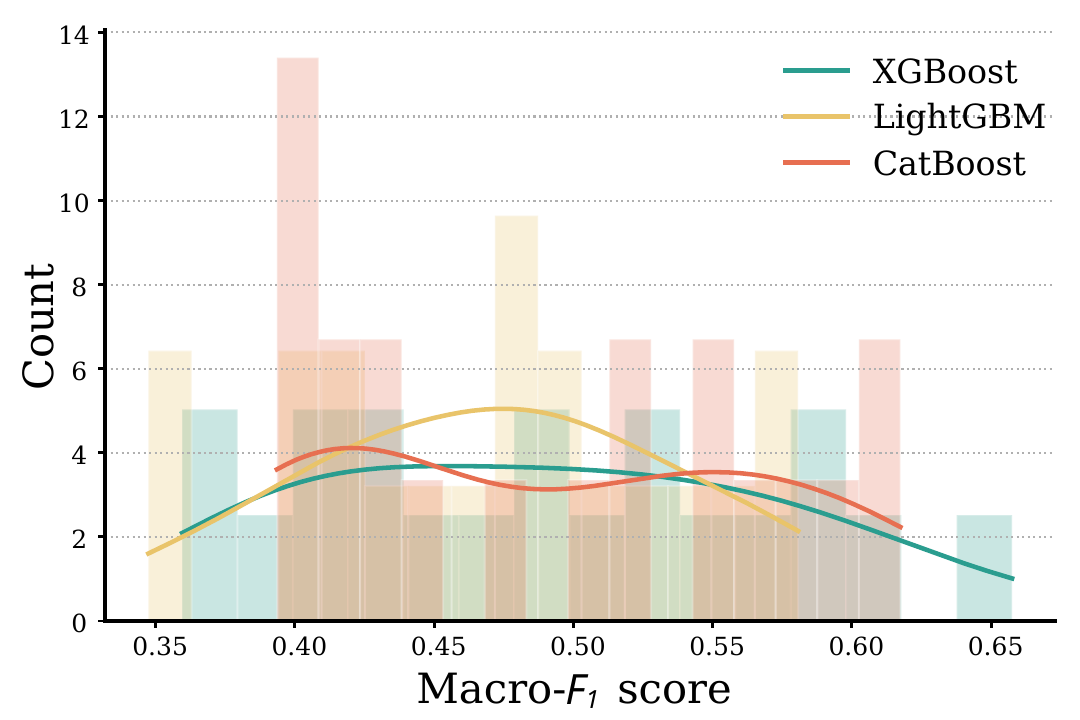}\\
    \includegraphics[width=0.8\linewidth]{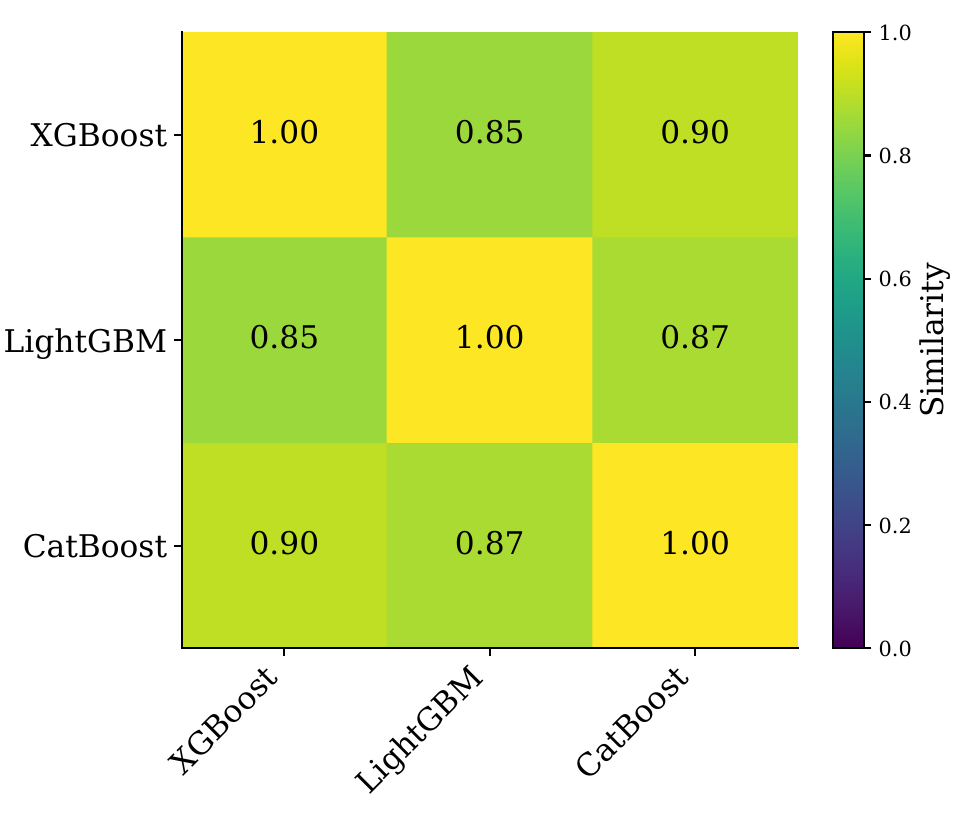}%
    \caption{Drug response classification. Top: Macro-$F_1$ scores across all drugs for different classifiers. Bottom: 
    Prediction agreement rate between considered models.}
    \label{fig:F1}
\end{figure}

\spara{Drug response classification.}
We employ gradient-boosting ensembles as our primary black-box predictor model: \textsc{XGBoost}, \textsc{CatBoost}, and \textsc{LightGBM}. These methods are well-suited to highly dimensional tabular data and can capture nonlinear effects and higher-order feature interactions, both of which are common in transcriptomic settings. 
Given the high dimensionality of transcriptomic data and the moderate sample size, we mitigate overfitting and ensure robust predictive performance by optimizing hyperparameters via stratified cross-validation on the training split, with particular emphasis on regularization and subsampling parameters, including the learning rate, tree depth, number of leaves, $\ell_1/\ell_2$ penalties, and feature and bagging fractions.  
Hyperparameter optimization is performed using Optuna \cite{akiba2019optuna}, an automatic hyperparameter search framework that employs adaptive sampling strategies to efficiently explore the parameter space.
After optimization, the model is retrained on the full training data with the best parameters and evaluated on the held-out test set.
Predictive performance is assessed through macro-F1 scores to account for the multiclass structure of the task. 
The top panel of \Cref{fig:F1} reports the distribution of test scores across drugs.\\
Among the evaluated models, we selected \textsc{LightGBM} for subsequent analyses because it exhibits the most stable performance across drugs and offers high computational efficiency and scalability, enabling rapid and consistent evaluation across multiple drugs.
Moreover, there is a strong agreement among the three considered model predictions, suggesting that all methods capture similar underlying structures in the data.
To quantify the prediction agreement rate between model outputs, we evaluated the fraction of instances for which each pair of models predicted the same class. Specifically, the similarity between two classifiers $A$ and $B$ is given by $\frac{1}{N} \sum_{i=1}^N \mathbb{1}[A(x_i) = B(x_i)]$. As shown at the bottom of \Cref{fig:F1}, the substantial agreement between all models indicates consistent predictions despite minor differences in classification performance.

\spara{Surrogate explainers training.}
\illora is implemented with low-rank linear layers, where the rank $r \ll m$ is fixed as the square root of the input dimensionality, i.e. $r = \sqrt{m}$. The latent space dimensionality is fixed to 96. For Jacobian penalty estimation~\cite{hoffman2019robust}, we employ 15 random projections. 
Training is done with Adam optimizer, fixing the learning rate to $10^{-3}$, with early stopping technique (patience of 30 epochs) to prevent overfitting.
To compute the loss functions, for input gene expressions and the projected latent vectors we use the cosine distance $d_{\mathit{cos}}(\mathbf{u},\mathbf{v}) = 1-\frac{\mathbf{u}\cdot\mathbf{v}}{\rvert\rvert\mathbf{u}\rvert\rvert ~\rvert\rvert\mathbf{v}\rvert\rvert}$. For the black-box score vectors, we employ the Euclidean distance. 
We consider as surrogate models: Logistic Regression to generate feature importance, and CART Decision Tree to derive rules.
For training the latter, we employ sparsity scheduling during training \illora, to reduce the number of input attributes (gene expressions) linearly combined in each latent feature to $2$.

Further details about the described steps are reported in Table~\ref{tab:drug_results}.
Since the surrogate models are trained using a one-vs-rest approach for the multi-class setting, the final fidelity is obtained by assigning the highest probability prediction across the class-specific surrogates.

\begin{table*}[htbp]
\centering
\scriptsize
\caption{
Per-drug dataset statistics after \BORUTA feature selection,
with performance metrics for LGBM (Macro-$F_1$) and surrogate
explainers (fidelity).
}
\label{tab:drug_results}

\setlength{\tabcolsep}{1.2mm}

\begin{tabular}{lll *{20}{c}}

\toprule

\multicolumn{2}{c}{}

& \rotatebox{90}{\drug{ Gefitinib}}
& \rotatebox{90}{\drug{ Navitoclax}}
& \rotatebox{90}{\drug{ Nutlin-3a (-)}}
& \rotatebox{90}{\drug{ Alisertib}}
& \rotatebox{90}{\drug{ BMS-536924}}
& \rotatebox{90}{\drug{ Erlotinib}}
& \rotatebox{90}{\drug{ Daporinad}}
& \rotatebox{90}{\drug{ Linsitinib}}
& \rotatebox{90}{\drug{ EPZ5676}}
& \rotatebox{90}{\drug{ RVX-208}}
& \rotatebox{90}{\drug{ LMB\_AB3}}
& \rotatebox{90}{\drug{ Ibrutinib}}
& \rotatebox{90}{\drug{ Venetoclax}}
& \rotatebox{90}{\drug{ ABT737}}
& \rotatebox{90}{\drug{ AZD3759}}
& \rotatebox{90}{\drug{ NVP-ADW742}}
& \rotatebox{90}{\drug{ WIKI4}}
& \rotatebox{90}{\drug{ Sepan.bromide}}
& \rotatebox{90}{\drug{ BMS-754807}}
& \rotatebox{90}{\drug{ SGC0946}} \\

\midrule

\multirowcell{2}{\textbf{Dataset}\\ \textbf{statistics}} &

\# Features
& 1337 & 1404 & 2774 & 2013 & 1250
& 1134 & 1408 & 2081 & 1767 & 1152
& 1447 & 1276 & 1305 & 1620 & 1197
& 1280 & 776 & 1452 & 1568 & 1308 \\

&\# Instances
& 586 & 591 & 592 & 579 & 586
& 583 & 321 & 591 & 587 & 546
& 454 & 543 & 586 & 585 & 590
& 586 & 585 & 586 & 578 & 580 \\

\midrule

\textbf{Black-box model} &
LGBM Macro-$F_1$
& 0.556 & 0.466 & 0.581 & 0.538 & 0.397
& 0.424 & 0.490 & 0.482 & 0.409 & 0.476
& 0.430 & 0.493 & 0.480 & 0.576 & 0.421
& 0.361 & 0.442 & 0.515 & 0.520 & 0.347 \\

\midrule

\multirowcell{2}{\textbf{Surrogates}\\ \textbf{fidelity}}&

Logistic Regression 
& 0.885 & 0.891 & 1.000 & 0.930 & 0.966
& 0.900 & 0.924 & 0.922 & 0.919 & 0.953
& 0.927 & 0.911 & 0.930 & 0.971 & 0.914
& 1.000 & 0.878 & 1.000 & 0.926 & 0.901 \\

&Decision Tree 
& 0.861 & 0.932 & 0.980 & 0.911 & 0.744
& 0.832 & 0.909 & 0.907 & 0.763 & 0.922
& 0.894 & 0.775 & 0.889 & 0.792 & 0.756
& 0.929 & 0.975 & 0.815 & 0.848 & 0.814 \\

\bottomrule

\end{tabular}

\end{table*}

\section{Ablation study on target discretization, feature selection, and black-box predictor}
\label{App:pipeline_sensitivity}

We evaluated the robustness of the proposed pipeline on eight representative GDSC drugs (\drug{Gefitinib}, \drug{Venetoclax}, \drug{Navitoclax}, \drug{Erlotinib}, \drug{Daporinad}, \drug{WIKI4}, \drug{BMS-536924}, and \drug{EPZ5676}) by varying one component at a time while keeping all others fixed. Specifically, we considered the following alternatives:

\begin{description}[leftmargin=*]
    \item[Label discretization.] We replaced the balanced tertile partitioning used in the main pipeline (33-33-33) with a more conservative quantile-based discretization, resulting in narrower extreme classes (25-50-25).

    \item[Regression objective.] We replaced the classification objective with a regression objective. To enable direct comparison with the main pipeline, predicted continuous values were subsequently discretized into tertiles using post-hoc binning.

    \item[Gradient-boosting (GB) feature selection.] We replaced \BORUTA-based feature selection with gradient-boosting feature ranking, selecting the same number of top-ranked features as identified by \BORUTA in the main pipeline.
    
    \item[Deep-learning predictor.] We replaced the proposed predictor with the state-of-the-art black-box classifier \textsc{TabPFN}~\cite{hollmann2025accurate}.
\end{description}

\begin{figure*}[t]
\setlength{\tabcolsep}{1pt}
\begin{tabular}{cccccc}
    \centering
    \includegraphics[height=0.105\linewidth]{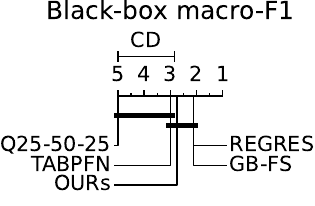} &  
    \includegraphics[height=0.105\linewidth]{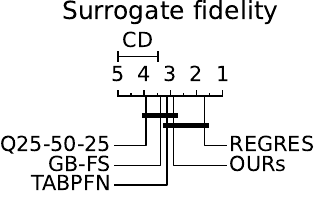} &
    \includegraphics[height=0.105\linewidth]{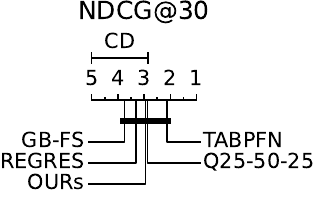} &
    \includegraphics[height=0.105\linewidth]{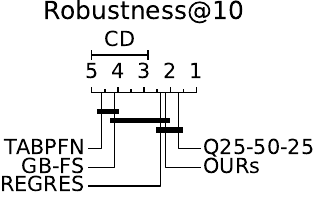} &
    \includegraphics[height=0.105\linewidth]{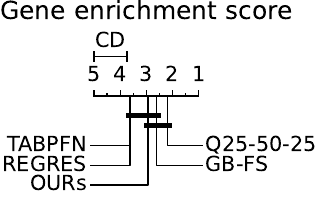} &
    \includegraphics[height=0.105\linewidth]{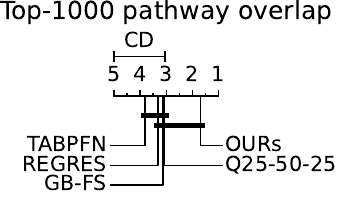}
\end{tabular}
    \caption{Pipeline sensitivity analysis. Aggregated performance of alternative end-to-end pipelines, assessed using CD diagram with Nemenyi at $\alpha=0.05$ of average ranks across eight GDSC drugs.}
\label{fig:ablation_pipeline}
\end{figure*}

Results are reported in Figure~\ref{fig:ablation_pipeline}, where statistical significance is assessed
using the Friedman test with Nemenyi post-hoc analysis~\cite{demsar2006statistical} at $\alpha = 0.05$. The analysis show that our pipeline with tertile-based splits (33-33-33) provides the best balance between performance and stability, suggesting that the preservation of balanced bins is more critical than the underlying training strategy. Moving to (25-50-25) splits reduces predictor performance and surrogate fidelity, confirming the importance of balanced classes for stable learning. The regression setting achieves marginally higher predictive accuracy and surrogate fidelity, but hinders interpretability. Replacing \BORUTA with GB feature selection, or \textsc{LightGBM} with TabPFN as black-box classifier, yields similar black-box predictive accuracy but reduces surrogate fidelity and feature importance robustness. Biological grounding is more sensitive: GB feature selection slightly favors (univariate) gene enrichment but degrades pathway overlap, while TabPFN shows reduced performance in both metrics.

\section{Decision rules: lengths and biological grounding}\label{app:decision_rules}

\begin{figure}[t]
    \centering
    \includegraphics[width=0.75\linewidth]{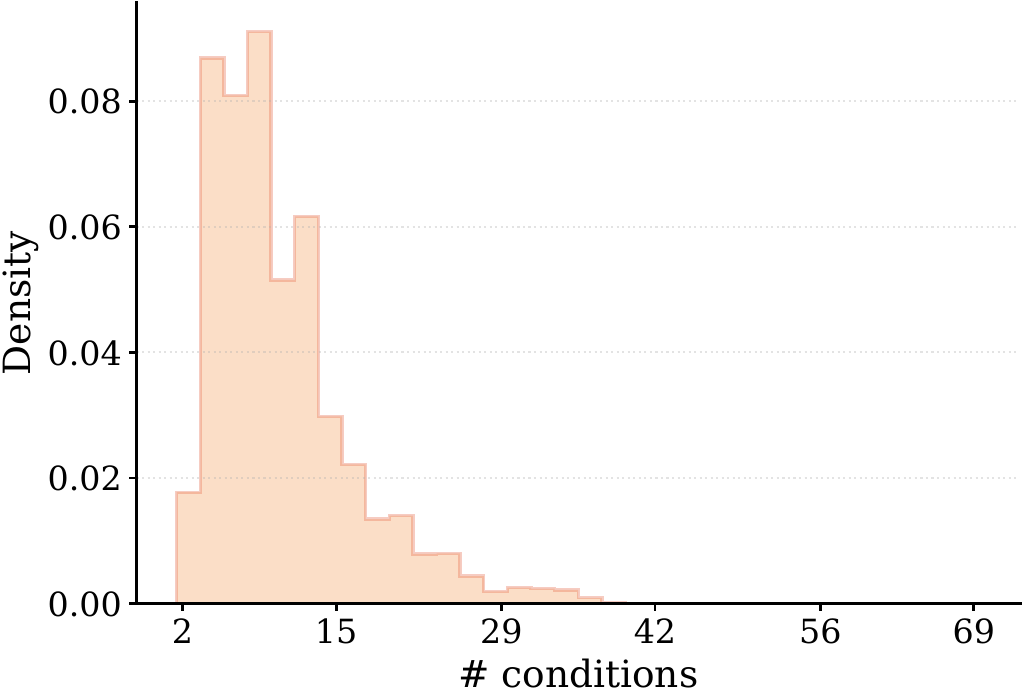} 
    \caption{Decision rules. Distribution of the length (i.e., number of conditions) of factual rules extracted via \illora.}
    \label{fig:rule_length_distribution}
\end{figure}

In Figure~\ref{fig:rule_length_distribution}, we show the distribution of rule lengths.
Moreover, in addition to the example rule reported in \Cref{sec:results}, \Cref{tab:pathway_rules_full} presents other example rules involving genes from MoA-related pathways for remaining drugs. 
\illora rules include genes directly associated to cancer resistance and tissue-specific phenotypes, some of which have been proposed as prognostic markers. 
Here we report some examples of how our rules connect to known and potential cancer markers.

\begin{table*}[h!]
\centering
\scriptsize
\setlength{\tabcolsep}{5pt}
\caption{
Complete set of pathway-aligned rules (labels restricted to low and high $IC_{50}$).
For each drug, we report the pathway whose genes most strongly align with the rule conditions.
The rule implication indicates the predicted response regime
(\textsc{Sensitive} = low $IC_{50}$, \textsc{Resistant} = high $IC_{50}$).
Pathway genes are highlighted in color.
}
\begin{tabular}{p{2.4cm} p{3.4cm} p{11.2cm}}
\toprule
\textbf{Drug} &
\textbf{Pathway} &
\textbf{Rule (pathway genes in color)} \\
\midrule
\drug{Alisertib} &
RNA Polymerase II Transcription &
$\pg{DDIT4}\!\leq\!-0.672 \wedge
 \pg{E2F1}\!>\!-0.008 \wedge
 \pg{ESRRG}\!>\!-0.223 \wedge
 \gene{KLK7}\!\leq\!2.568 \wedge
 \gene{KRT27}\!>\!-0.205 \wedge
 \gene{P4HTM}\!>\!-0.277
 \;\Rightarrow\; \textsc{Resistant}$
\\

\drug{Erlotinib} &
Nervous system development &
$\pg{CNTNAP1}\!>\!-0.891 \wedge
 \pg{EVL}\!\leq\!-1.395 \wedge
 \gene{FAM214B}\!\leq\!0.212 \wedge
 \gene{IL13RA1}\!\leq\!0.976 \wedge
 \gene{IRF6}\!>\!0.390 \wedge
 \pg{PSEN2}\!>\!0.133
 \;\Rightarrow\; \textsc{Resistant}$ \\

\drug{Linsitinib} &
Signaling by Receptor Tyrosine Kinases &
$\gene{CGN}\!\leq\!0.913 \wedge
 \gene{FAM83C}\!>\!-0.733 \wedge
 \pg{GALNT3}\!>\!1.121 \wedge
 \pg{GRAP2}\!>\!-1.437 \wedge
 \gene{LYPD1}\!>\!-2.418 \wedge
 \pg{POLR2D}\!\leq\!0.297 \wedge
 \gene{SLC10A6}\!>\!-0.530 \wedge
 \pg{TNS4}\!\leq\!3.993
 \;\Rightarrow\; \textsc{Sensitive}$ \\

\drug{NVP-ADW742} &
Infectious disease &
$\pg{CD8B}\!\leq\!-0.891 \wedge
 \pg{DOCK1}\!\leq\!3.386 \wedge
 \pg{GHRH}\!>\!-2.722 \wedge
 \gene{LHX2}\!\leq\!2.241 \wedge
 \gene{PPIC}\!\leq\!4.478 \wedge
 \gene{UBE2J1}\!>\!0.824 \wedge
 \pg{WAS}\!\leq\!3.977 \wedge
 \gene{ZNF629}\!\leq\!1.287
 \;\Rightarrow\; \textsc{Resistant}$ \\

\drug{Daporinad} &
RNA Polymerase II Transcription &
$\gene{APMAP}\!>\!-1.699 \wedge
 \pg{AURKA}\!\leq\!5.094 \wedge
 \gene{KIFC1}\!\leq\!3.194 \wedge
 \gene{KRT1}\!\leq\!0.645 \wedge
 \gene{LCE3E}\!\leq\!0.344 \wedge
 \pg{MAF}\!\leq\!2.123 \wedge
 \gene{MUC6}\!\leq\!1.270 \wedge
 \pg{RHEB}\!>\!-0.258 \wedge
 \pg{RNMT}\!\leq\!0.862 \wedge
 \gene{THAP11}\!\leq\!1.177 \wedge
 \pg{UPF3B}\!\leq\!1.168 \wedge
 \gene{VSIG8}\!\leq\!0.635
 \;\Rightarrow\; \textsc{Resistant}$ \\

\drug{Ibrutinib} &
Cytokine Signaling in Immune system &
$\pg{ANXA2}\!>\!-3.612 \wedge
 \gene{CD8B}\!\leq\!4.818 \wedge
 \gene{CDH3}\!>\!0.633 \wedge
 \pg{FASLG}\!\leq\!0.551 \wedge
 \gene{FCRL6}\!\leq\!0.838 \wedge
 \pg{IRF6}\!\leq\!7.289 \wedge
 \gene{LAG3}\!>\!-3.951 \wedge
 \pg{LTA}\!\leq\!2.004 \wedge
 \pg{PSMB11}\!>\!-0.076 \wedge
 \gene{SLAMF8}\!>\!-2.508 \wedge
 \gene{SPRR1A}\!>\!-0.505 \wedge
 \gene{WDR91}\!>\!-2.491 \wedge
 \gene{WSB2}\!>\!-2.868
 \;\Rightarrow\; \textsc{Sensitive}$ \\

\drug{Navitoclax} &
Signaling by Nuclear Receptors &
$\pg{CAV1}\!\leq\!1.255 \wedge
 \gene{CLEC17A}\!\leq\!0.214 \wedge
 \gene{CLK2}\!\leq\!0.570 \wedge
 \gene{FBF1}\!\leq\!1.463 \wedge
 \gene{HBB}\!>\!-4.392 \wedge
 \pg{MOV10}\!\leq\!0.973 \wedge
 \gene{S100G}\!>\!-0.180 \wedge
 \pg{ZDHHC7}\!\leq\!1.426
 \;\Rightarrow\; \textsc{Sensitive}$ \\

\drug{Sepan. bromide} &
Cell Cycle Checkpoints &
$\pg{BIRC5}\!>\!-0.159 \wedge
 \gene{CXorf21}\!\leq\!-0.244 \wedge
 \gene{GLYATL3}\!\leq\!0.492 \wedge
 \gene{MSH6}\!>\!-0.062 \wedge
 \pg{PCBP4}\!\leq\!-0.349 \wedge
 \gene{ST8SIA3}\!\leq\!0.673
 \;\Rightarrow\; \textsc{Sensitive}$ \\

\drug{WIKI4} &
Intracellular signaling by second messengers &
$\gene{LCE1A}\!\leq\!-0.097 \wedge
 \gene{NFKB2}\!\leq\!1.117 \wedge
 \pg{PHC2}\!>\!-0.791 \wedge
 \pg{PSMB11}\!\leq\!0.058 \wedge
 \gene{TMEM102}\!>\!-1.832 \wedge
 \gene{TSACC}\!\leq\!0.410
 \;\Rightarrow\; \textsc{Sensitive}$ \\

\drug{BMS-754807} &
Signaling by Nuclear Receptors &
$\gene{ANKRD66}\!>\!-0.017 \wedge
 \gene{ARSH}\!\leq\!0.098 \wedge
 \gene{CYB5R4}\!>\!0.337 \wedge
 \pg{PCK1}\!>\!-0.567 \wedge
 \pg{PRKCZ}\!\leq\!1.330 \wedge
 \gene{RPS8}\!\leq\!0.706 \wedge
 \pg{SMC3}\!\leq\!-0.298 \wedge
 \gene{UQCRH}\!>\!-0.263 \wedge
 \gene{WDR26}\!>\!-0.844 \wedge
 \gene{ZNF567}\!\leq\!0.113
 \;\Rightarrow\; \textsc{Resistant}$ \\

\drug{RVX-208} &
Infectious disease &
$\pg{CD28}\!>\!-0.781 \wedge
 \gene{EXPH5}\!\leq\!-0.265 \wedge
 \pg{GNG8}\!\leq\!-0.226 \wedge
 \pg{GNG8}\!>\!-0.949 \wedge
 \gene{IL21R}\!\leq\!1.123 \wedge
 \gene{KIR3DL2}\!>\!-0.498 \wedge
 \gene{NAV1}\!>\!-3.954 \wedge
 \pg{PSME3}\!>\!-0.498 \wedge
 \gene{STATH}\!\leq\!0.453 \wedge
 \gene{STATH}\!>\!-0.029
 \;\Rightarrow\; \textsc{Sensitive}$ \\

\drug{Nutlin-3a (-)} &
Cellular responses to stress &
$\gene{BOD1}\!\leq\!3.477 \wedge
 \pg{CDKN1A}\!\leq\!8.755 \wedge
 \gene{ETV3}\!>\!-2.026 \wedge
 \gene{HOXD13}\!\leq\!2.154 \wedge
 \gene{ITPRIPL1}\!\leq\!2.021 \wedge
 \gene{OR5H14}\!>\!-0.012 \wedge
 \gene{OR9Q1}\!>\!-0.010 \wedge
 \gene{PDE1B}\!>\!-0.391 \wedge
 \gene{PPIL1}\!>\!-1.377 \wedge
 \pg{SH3BP4}\!>\!-0.076 \wedge
 \gene{TBC1D31}\!\leq\!1.742 \wedge
 \pg{TERF2IP}\!\leq\!2.249 \wedge
 \pg{TUBA4B}\!\leq\!0.373 \wedge
 \gene{ZNF705G}\!>\!-0.149
 \;\Rightarrow\; \textsc{Sensitive}$ \\

\drug{BMS-536924} &
Infectious disease &
$\pg{CEBPD}\!>\!0.120 \wedge
 \gene{CPNE5}\!>\!-0.712 \wedge
 \gene{GNRH2}\!>\!-0.830 \wedge
 \gene{HMGB1}\!\leq\!0.924 \wedge
 \gene{KRTAP22-1}\!>\!-0.008 \wedge
 \pg{LCK}\!>\!-2.916 \wedge
 \gene{MAPK6}\!\leq\!0.491 \wedge
 \gene{NMS}\!\leq\!0.103 \wedge
 \gene{OR51D1}\!\leq\!0.009 \wedge
 \pg{POLR2J}\!\leq\!0.015 \wedge
 \gene{RASAL3}\!\leq\!1.308
 \;\Rightarrow\; \textsc{Sensitive}$ \\

\drug{ABT737} &
Signaling by Nuclear Receptors &
$\gene{EPHA2}\!\leq\!1.960 \wedge
 \pg{GNG12}\!\leq\!3.080 \wedge
 \gene{IL21}\!\leq\!0.286 \wedge
 \gene{RILPL1}\!\leq\!0.702
 \;\Rightarrow\; \textsc{Sensitive}$ \\

\drug{Gefitinib} &
Signaling by Receptor Tyrosine Kinases &
$\gene{DEFB114}\!\leq\!0.061 \wedge
 \pg{FGF16}\!\leq\!0.548 \wedge
 \pg{FGFBP1}\!\leq\!6.297 \wedge
 \gene{GBF1}\!>\!-1.653 \wedge
 \gene{GDF10}\!>\!-2.337 \wedge
 \gene{GIMAP7}\!\leq\!-0.128 \wedge
 \gene{RPS12}\!>\!-1.344 \wedge
 \gene{ZNF705D}\!\leq\!0.141
 \;\Rightarrow\; \textsc{Resistant}$ \\

\drug{Venetoclax} &
Apoptosis &
$\gene{BOD1L2}\!>\!0.144 \wedge
 \gene{CAPNS2}\!\leq\!1.990 \wedge
 \gene{CDKN3}\!\leq\!1.766 \wedge
 \gene{CPOX}\!>\!-2.232 \wedge
 \gene{DNAJB5}\!\leq\!1.264 \wedge
 \gene{MOB3B}\!\leq\!0.058 \wedge
 \gene{MOB3B}\!>\!-5.296 \wedge
 \pg{PSMA8}\!\leq\!0.556 \wedge
 \pg{TJP1}\!>\!-0.689 \wedge
 \gene{UBA1}\!>\!-0.100
 \;\Rightarrow\; \textsc{Sensitive}$ \\

\drug{AZD3759} &
MAPK family signaling cascades &
$\gene{ADAT3}\!\leq\!0.679 \wedge
 \gene{CD226}\!>\!0.058 \wedge
 \gene{CUEDC1}\!\leq\!1.029 \wedge
 \gene{CUEDC1}\!>\!-5.662 \wedge
 \gene{DPYSL3}\!>\!-6.229 \wedge
 \pg{EGFR}\!\leq\!-0.968 \wedge
 \gene{EPPK1}\!\leq\!3.159 \wedge
 \gene{OR6K3}\!\leq\!-0.009 \wedge
 \gene{OR6K3}\!>\!-0.068 \wedge
 \gene{PLEC}\!\leq\!0.413 \wedge
 \gene{PLEC}\!>\!-1.351 \wedge
 \pg{PTPN7}\!\leq\!-0.127 \wedge
 \pg{RGL3}\!\leq\!0.241 \wedge
 \pg{RGL3}\!>\!-2.799 \wedge
 \gene{SOX15}\!>\!-1.880 \wedge
 \gene{ZNF80}\!\leq\!0.080
 \;\Rightarrow\; \textsc{Resistant}$ \\

\drug{SGC0946} &
Chromatin modifying enzymes &
$\gene{CASP14}\!\leq\!2.049 \wedge
 \gene{CBR3}\!>\!0.624 \wedge
 \gene{CXCR6}\!>\!-2.422 \wedge
 \gene{GGCT}\!>\!0.017 \wedge
 \pg{PAX3}\!\leq\!1.078 \wedge
 \gene{PLA2G7}\!>\!0.192
 \;\Rightarrow\; \textsc{Sensitive}$ \\

\drug{EPZ5676} &
Chromatin modifying enzymes &
$\gene{GGT6}\!\leq\!3.291 \wedge
 \gene{GGT6}\!>\!-1.992 \wedge
 \gene{HOXD10}\!\leq\!8.914 \wedge
 \gene{IFNA2}\!>\!-0.130 \wedge
 \gene{IL1F10}\!\leq\!0.403 \wedge
 \gene{IL1F10}\!>\!-0.177 \wedge
 \pg{KDM2B}\!>\!-4.287 \wedge
 \gene{KRT25}\!\leq\!0.282 \wedge
 \gene{LALBA}\!\leq\!0.572 \wedge
 \gene{LALBA}\!>\!-2.288 \wedge
 \gene{MCEMP1}\!\leq\!0.319 \wedge
 \gene{OR2T35}\!\leq\!0.047 \wedge
 \gene{OR2Z1}\!\leq\!0.075 \wedge
 \gene{OR2Z1}\!>\!-0.460 \wedge
 \gene{OR51D1}\!>\!-0.012 \wedge
 \gene{PSMB11}\!>\!-0.169 \wedge
 \gene{RPF1}\!\leq\!3.205 \wedge
 \pg{SMARCA2}\!\leq\!0.828 \wedge
 \gene{SSTR3}\!\leq\!0.761 \wedge
 \gene{TEX37}\!\leq\!0.483 \wedge
 \gene{TIMD4}\!>\!-2.673 \wedge
 \gene{TRIM26}\!\leq\!-0.245 \wedge
 \gene{ZNF440}\!\leq\!0.564 \wedge
 \gene{ZNF705D}\!>\!-0.256 \wedge
 \gene{ZP2}\!>\!-0.935
 \;\Rightarrow\; \textsc{Sensitive}$ \\

\bottomrule
\end{tabular}
\label{tab:pathway_rules_full}
\end{table*}

\drug{Erlotinib} is a receptor tyrosine kinase inhibitor (RTKI) targeting the Epidermal Growth Factor Receptor (EGFR). Increased Notch signalling has been implicated in Erlotinib resistance~\cite{xie2013notch, mur2020notch}. Accordingly, high levels of PSEN2, a core component of the $\gamma$-secretase complex required for Notch activation, were associated with resistance in our model. Other contributions are more nuanced: IRF6~\cite{muralidharan2024breast, botti2011developmental} and IL13RA1 ~\cite{bednarz2020interleukins, shi2022loss} have tumor-dependent roles. Consistent with our rule, reduced EVL levels have been associated to colorectal cancer ~\cite{yu2023elevated} and increased metastasis in breast cancer~\cite{padilla2018actin}. Other genes have not been thoroughly investigated in cancer, but could be potential biomarkers: among them, CTNAP1 has been suggested as a clear cell renal cell carcinoma marker in an independent bioinformatic analysis~\cite{li2022m2}.

\drug{Gefitinib} is also a EGR-RTKI. Among the genes associated to Gefitinib resistance, GIMAP7 is significantly reduced across cancers, specifically at advanced stages ~\cite{qin2022gimap7}, and low levels sustains cell proliferation and reduce apoptosis in lung adenocarcinoma~\cite{cui2024gimap7}. GBF1, FGFBP1, RPS12 have cancer type-specific effects on survival~\cite{uhlen2015tissue}, and their interactions have not been addressed to the best of our knowledge. The rule also include new genes such as DEFB114.

\drug{Navitoclax} triggers apoptosis by inhibiting BCL-2, BCL-xL and BCL-w~\cite{tse2008abt}. Lower CAV1 is consistent with reduced survival signaling and increased apoptotic susceptibility~\cite{kabir2025inhibition}, consistent with our predicted higher susceptibility to Navitoclax. MOV10 downregulation increases levels of the INK4a tumor suppressor~\cite{messaoudi2010role}, and CLK2 inhibition increases apoptosis~\cite{MUCKA20231303}, consistent with their association to sensitivity. Other genes are involved in stress response and calcium influx, both linked to apoptosis. 

Overall, our rules identify cancer-associated and additional genes, generating quantitative hypotheses for experimental validation.

\section{Gene-gene ranking: supplementary  results}
\label{app: Biological relevance of gene-gene ranking}

\begin{figure*}[t]
    \centering
    \setlength{\tabcolsep}{2pt}

    \begin{tabular}{cccc}
        \includegraphics[trim={0 7.5mm 0 0}, clip,width=0.245\linewidth]{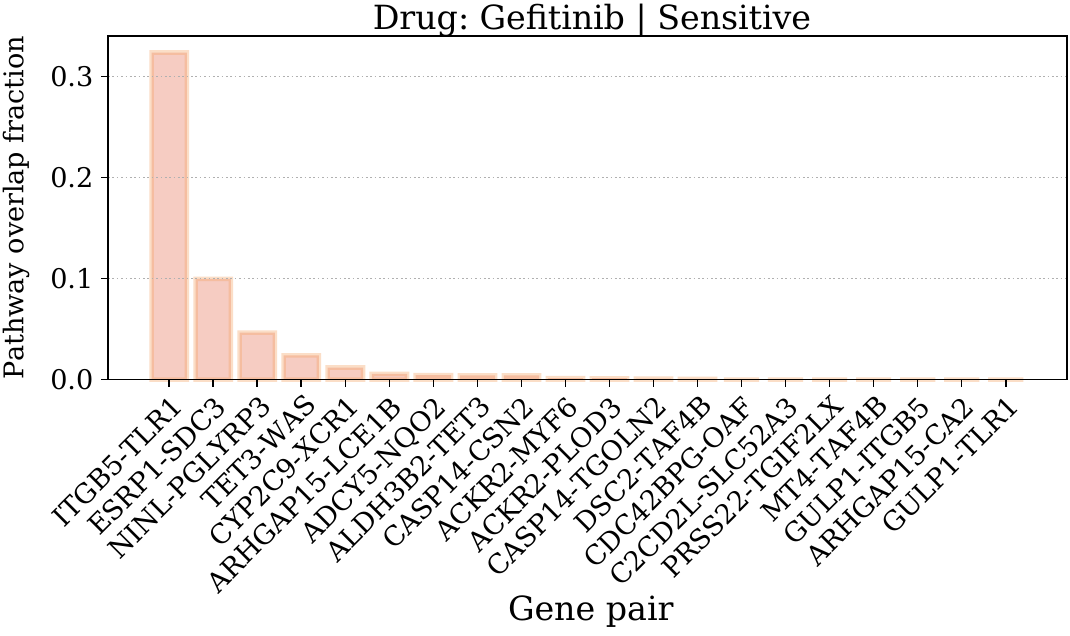} &
        \includegraphics[trim={0 7.5mm 0 0}, clip,width=0.245\linewidth]{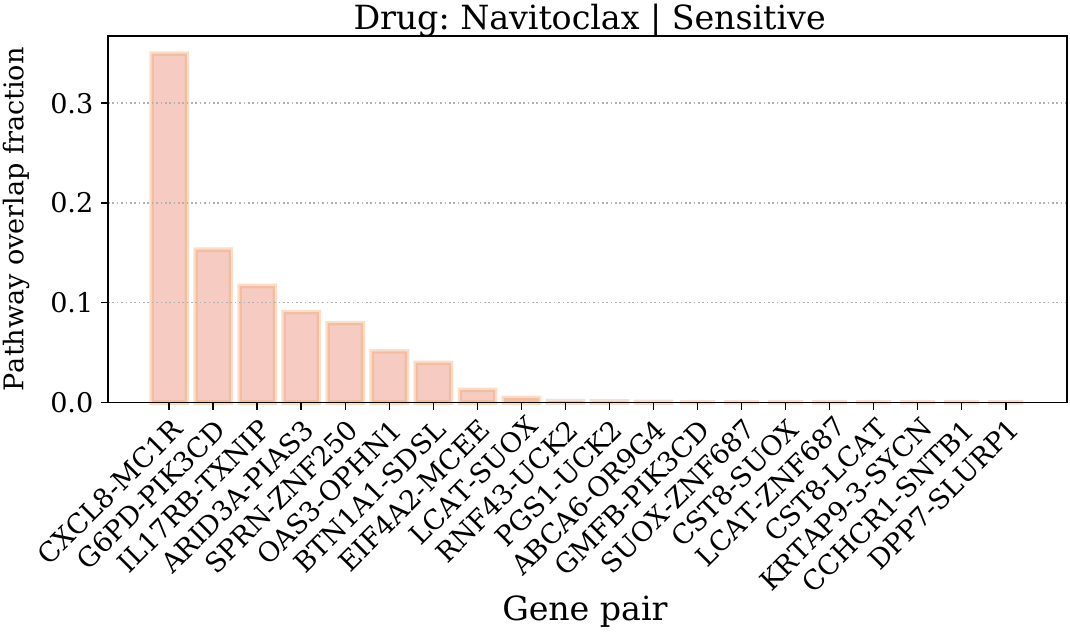} &
        \includegraphics[trim={0 7.5mm 0 0}, clip,width=0.245\linewidth]{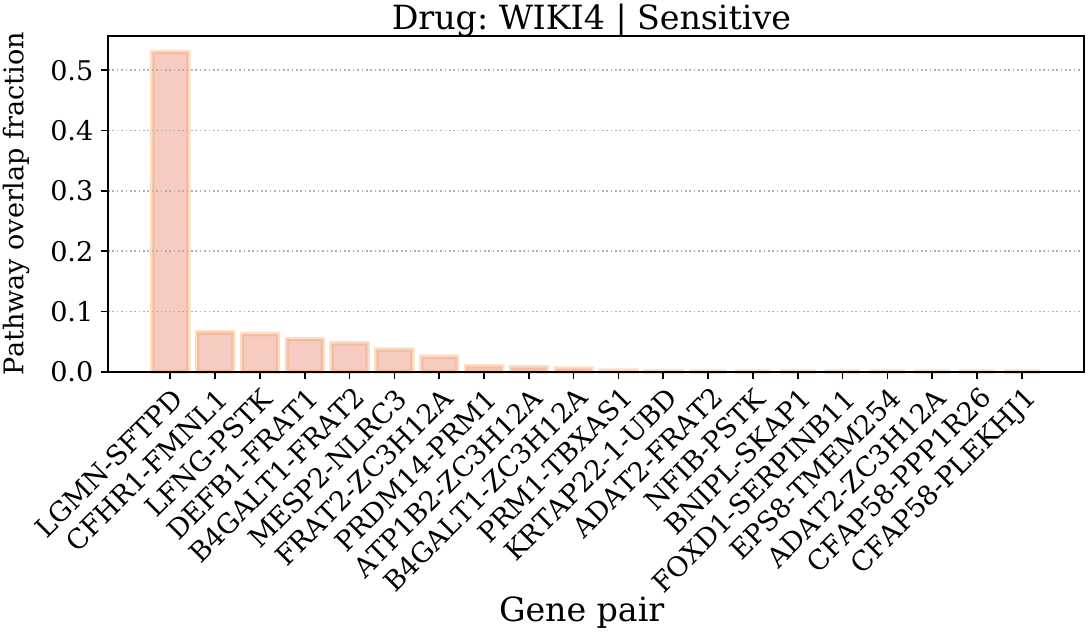} &
        \includegraphics[trim={0 7.5mm 0 0}, clip,width=0.245\linewidth]{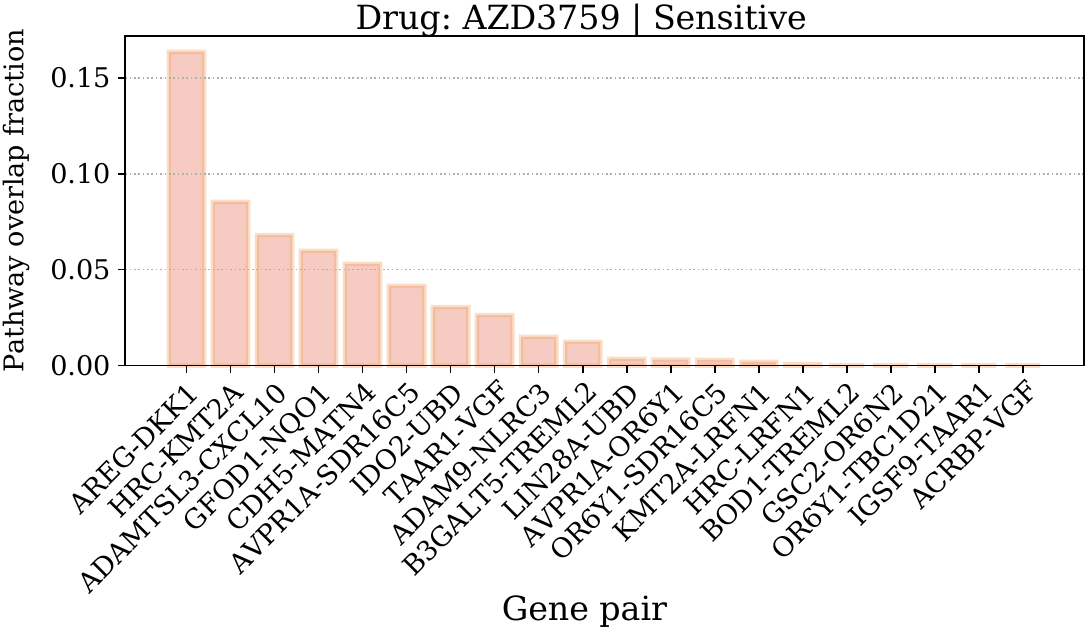}\\
        \includegraphics[width=0.245\linewidth]{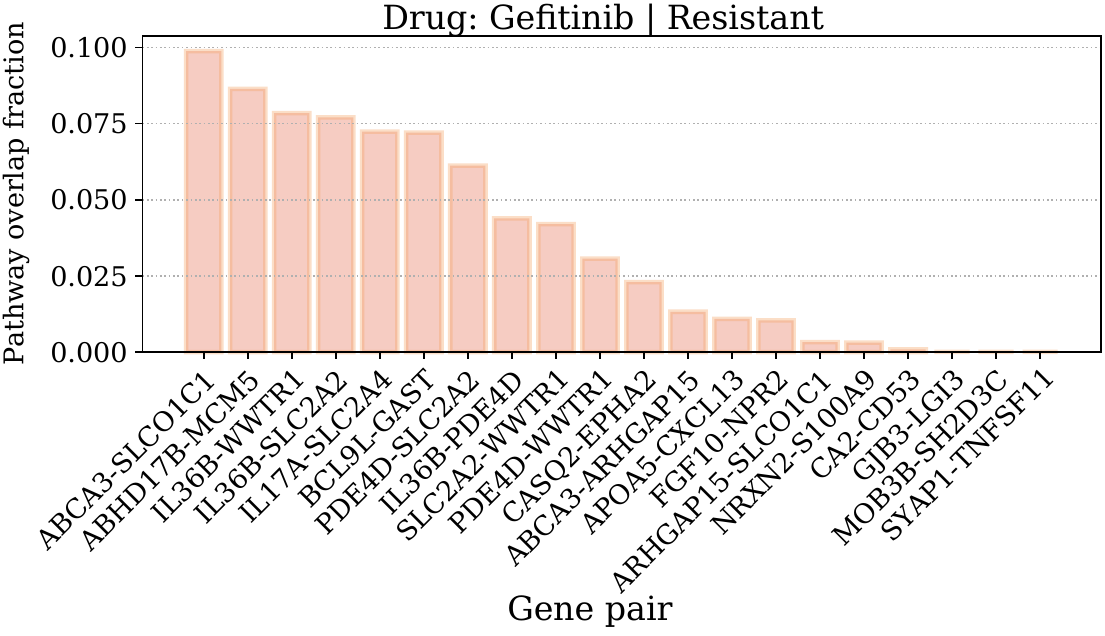} &
        \includegraphics[width=0.245\linewidth]{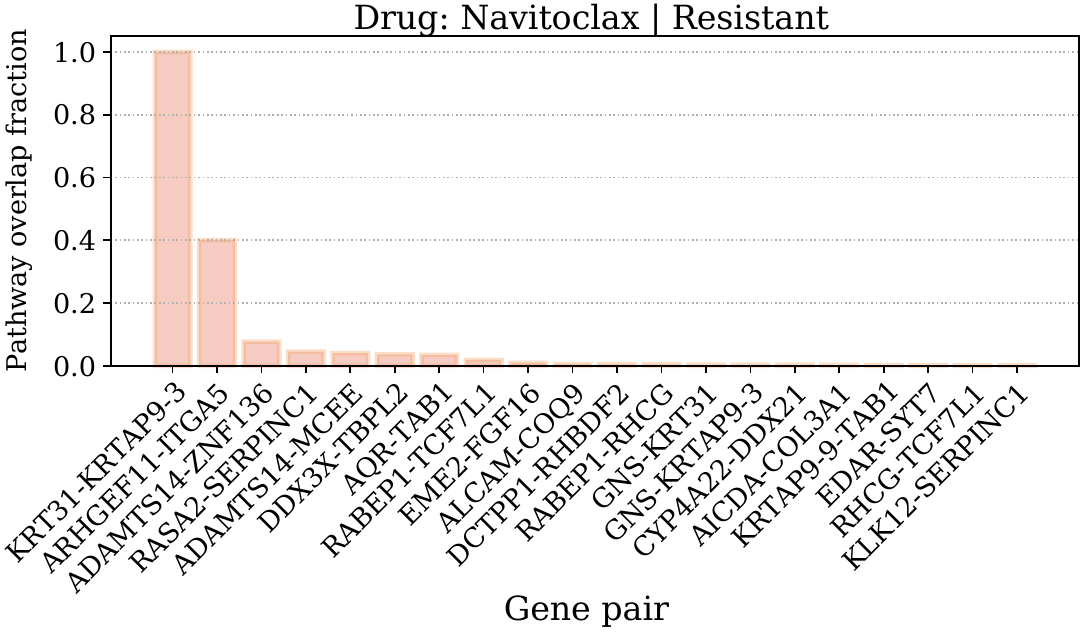} &
        \includegraphics[width=0.245\linewidth]{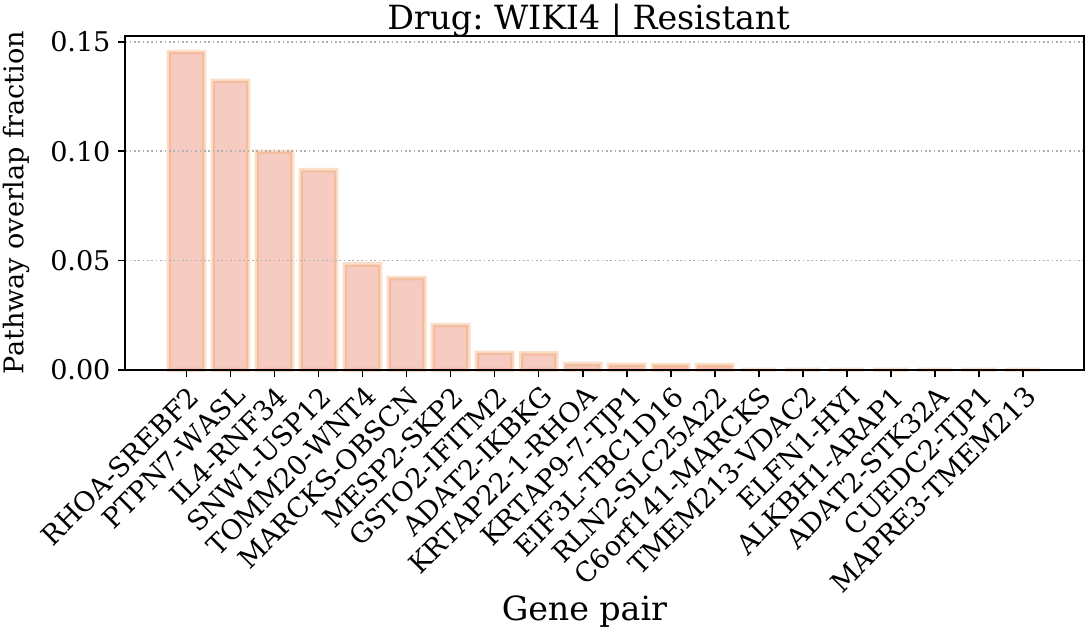} &
        \includegraphics[width=0.245\linewidth]{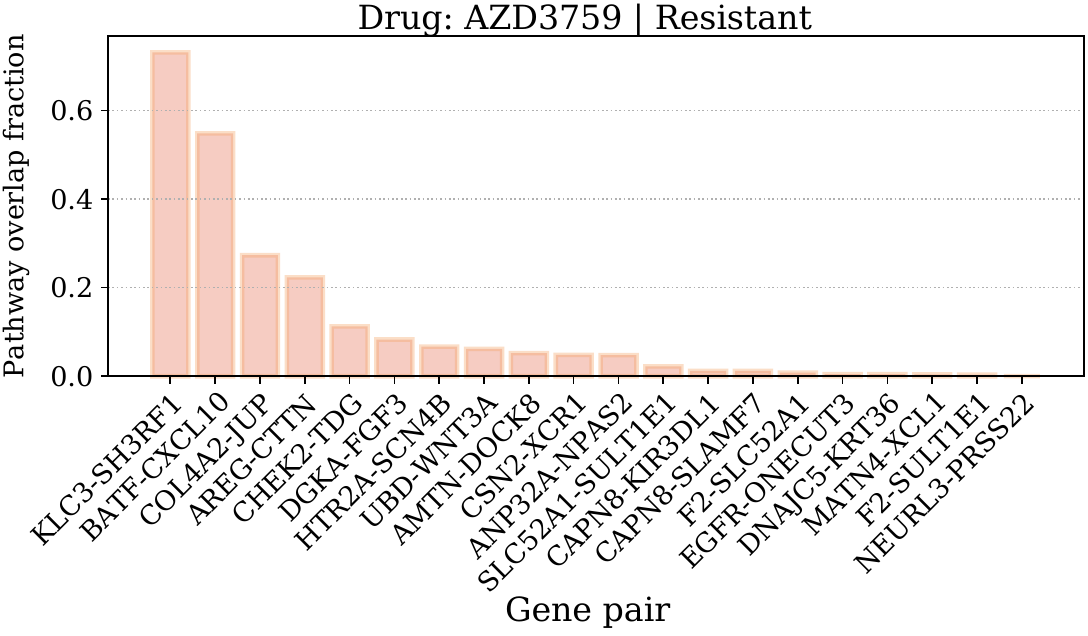}
    \end{tabular}

    \caption{Biological relevance.
    Pathway Overlap values for the top-20 gene pairs in four exemplifying drugs (\drug{Gefitinib}, \drug{Navitoclax}, \drug{WIKI4}, and \drug{AZD3759}).
    Top row: \emph{Sensitive} class.
    Bottom row: \emph{Resistant} class. 
    }
    \label{Appfig:pathway_colored_top20_5drugs_pairwise} 
\end{figure*}

\spara{Biological relevance.} 
Extending the analysis presented in the main text, in Figure~\ref{Appfig:pathway_colored_top20_5drugs_pairwise} we illustrate the 20 gene pairs with the largest PO among the 50 with the largest lift in four exemplifying drugs.
For \drug{Navitoclax}-sensitive cell lines, highly ranked gene pairs are enriched in pathways related to immune processes and carbon metabolism in cancer, the latter being a known determinant of response~\cite{al2018targeting}. 
In \drug{Gefitinib}, top interactions encompass more general signal transduction and cell-cell communication pathways, which are significantly enriched among gene pairs associated with resistance. \\
For \drug{WIKI4}, a potent inhibitor of Wnt signalling (it hinders tankyrase, leading to inhibition of AXIN degradation and suppression of Wnt-activated transcription), among the top 10 most important gene pairs associated to sensitive cell lines, the 3 genes with the largest number of interactions (4-6) have strong links to Wnt signaling: FRAT2 promotes Wnt activity and cell sensitization~\cite{New1}, B4GALT1 regulates the Wnt axis~\cite{New2}, and ZC3H12A negatively modulates Wnt signalling~\cite{New3}. The ZC3H12A-FRAT2 pair would thus lead to higher Wnt dependence, consistent with increased WIKI4 sensitivity, while B4GALT interactions further modulate Wnt activity and responsiveness.\\
In \drug{AZD3759}, a reversible EGFR tyrosine kinase inhibitor effective in cell lines with strong EGFR activation~\cite{New4},
top pairs include: (a) LIN28A-UBD, which suppresses let-7 miRNA biogenesis, leading to EGFR upregulation~\cite{New5} and UBD promotes tumor progression and invasion~\cite{New6,New7}, and can induce resistance by EGFR ubiquitination~\cite{New8}; and (b) RASSF9-REG4, both central to EGFR signalling: RASSF9 activation leads to upregulation of the MEK-ERK pathway downstream of EGFR, while REG4 acts as an upstream EGFR activator. Moreover, VGF (in 2 pairs) has been directly linked to EGFR inhibitor resistance in lung cancer~\cite{New9,New10}.\\
These observations further suggest that highly ranked gene pairs consistently recover biologically grounded processes across drugs, supporting the robustness of our approach and reducing the likelihood that the observed associations arise from spurious correlations.

\begin{figure}[t]
    \centering
    \includegraphics[width=\linewidth]{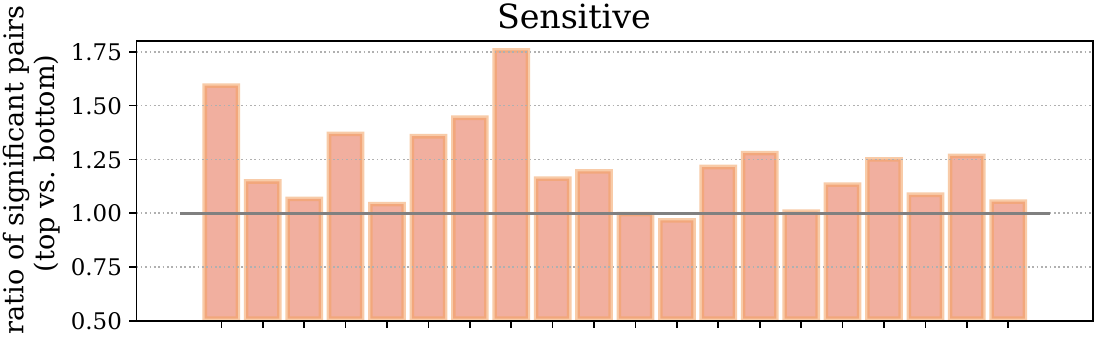}\\ 
    \includegraphics[width=\linewidth]{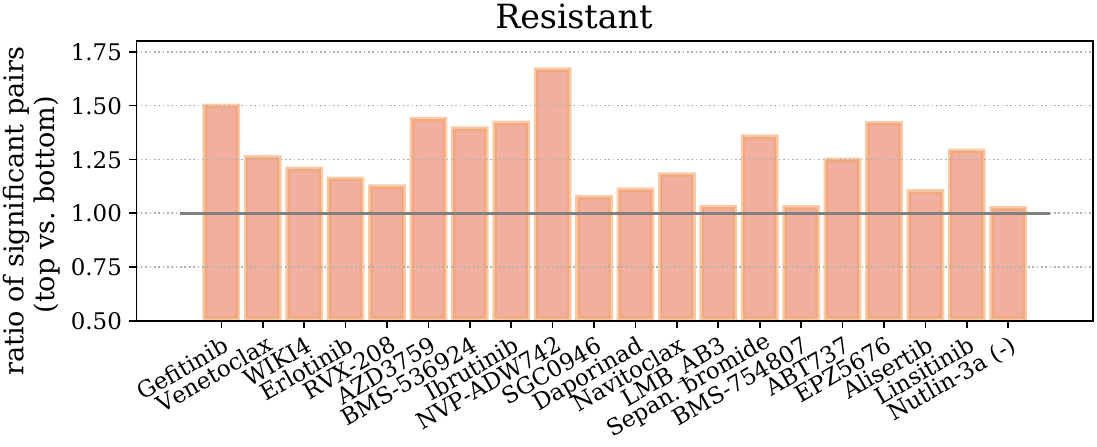}
    \caption{Task-based relevance. Proportion of significant interactions in top-1,000 ranked gene pairs compared to bottom-1,000 ranked ones for multiple drugs. }
\label{fig:pairwise_logistisic_pvalues_sensitive}
\end{figure}

\spara{Task-based relevance.} To ensure that highly ranked gene pairs capture task-relevant signals, we test whether the interaction ranking produced by \illora is consistent with statistical evidence derived from the same prediction task, as described in section \ref{subsec:beyond univariate}. 
Specifically, in this analysis we compare the top 1,000 ranked gene pairs with the bottom 1,000; additional random samples are then compared to the top 1,000 as a robustness check, 
yielding consistent results.
For each pair of genes $(g_i, g_j)$ we fit a logistic regression model $\tilde{y}_{ij} = \beta_i x_i + \beta_j x_j + \beta_{ij} x_i x_j$, including their interaction term $x_i x_j$,  and evaluate its significance via the Wald test~\cite{agresti2015foundations}. \Cref{fig:pairwise_logistisic_pvalues_sensitive} reports the ratio of number of pairs with \textit{p}-value below $0.05$ for the top and bottom gene pairs in the ranking, showing that up to  $70$–$75\%$ of top-ranked interactions are significantly higher than for bottom pairs.

\begin{figure}[t]
    \centering
    \begin{tabular}{c@{\hspace{0.4cm}}|@{\hspace{0.4cm}}c}
    \includegraphics[width=0.35\linewidth]{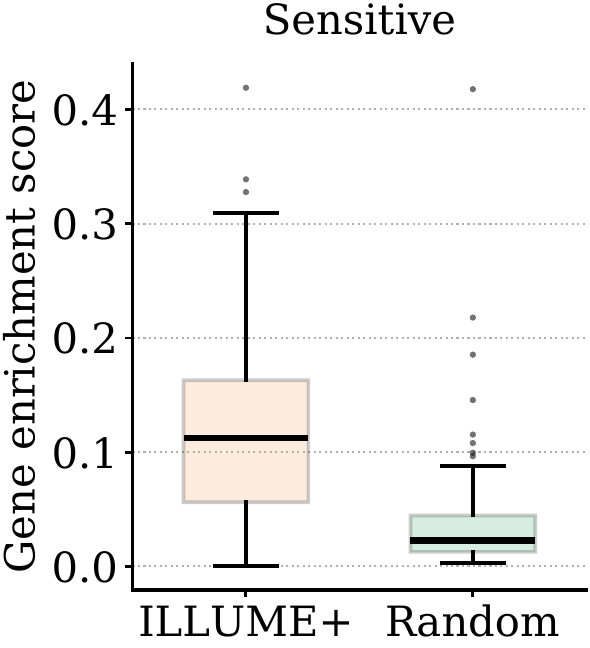} 
    \hspace{0.25cm}
    &
    \hspace{0.25cm}
    \includegraphics[width=0.35\linewidth]{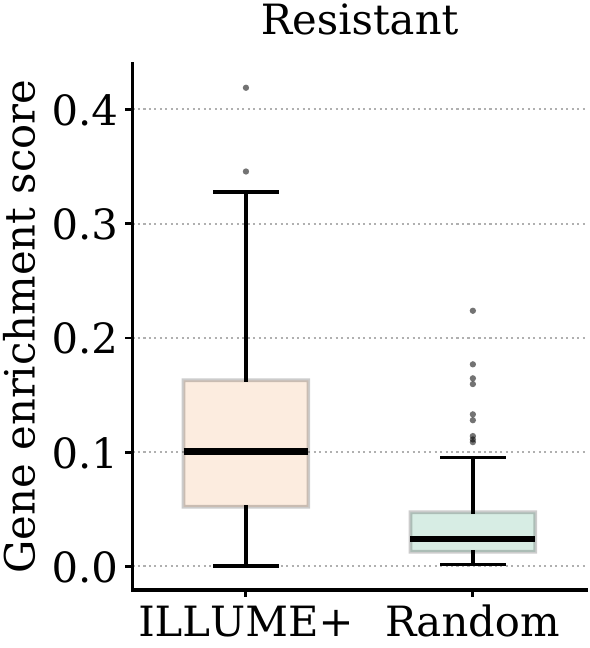} 
    \end{tabular}
    \caption{Distribution of bivariate enrichment scores for the rankings of pairwise interactions given by \illora and a random baseline.}
    \label{fig:pairwise_enrichment_score}
\end{figure}

\spara{Bivariate gene enrichment.}
In line with unweighted enrichment score for single-gene ranks~\cite{subramanian2005gene}, we define bivariate enrichment for gene-gene pair ranks as follows:
\[
\mathrm{biES}
=
\max_{1 \le k \le M}
\left|
\sum_{h=1}^{k}
\left(
\frac{\mathbb{1}\!\left[\pi_h \in \mathcal{S}_{\mathrm{pair}}\right]}{|\mathcal{S}_{\mathrm{pair}}|}
-
\frac{\mathbb{1}\!\left[\pi_h \notin \mathcal{S}_{\mathrm{pair}}\right]}{M-|\mathcal{S}_{\mathrm{pair}}|}
\right)
\right|,
~~~
M=\binom{m}{2},
\]
where $\{\pi_h\}_{h=1}^M$ is the ranked list of gene pairs (each $\pi_h = (g_{i_h}, g_{j_h})$)
and $\mathcal{S}_{\text{pair}}$ denotes the set of relevant gene pairs co-occurring on known MoA-related pathways.
\Cref{fig:pairwise_enrichment_score} presents bivariate gene enrichment scores for gene-gene rankings produced by \illora, compared against a random baseline where relevance labels are permuted. Consistent with the univariate analysis, pairwise rankings from \illora are more likely to overrepresent known biological pathways, where ranked gene pairs co-occur—than those obtained by random ranking.

\begin{figure*}
\centering
\includegraphics[trim={0 0 5mm 0}, clip, width=0.85\linewidth]{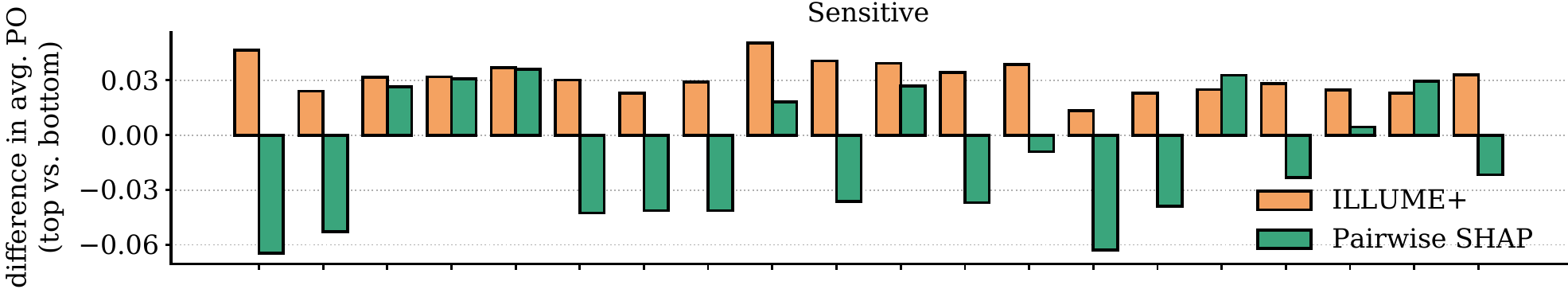} \\
\includegraphics[width=0.85\linewidth]{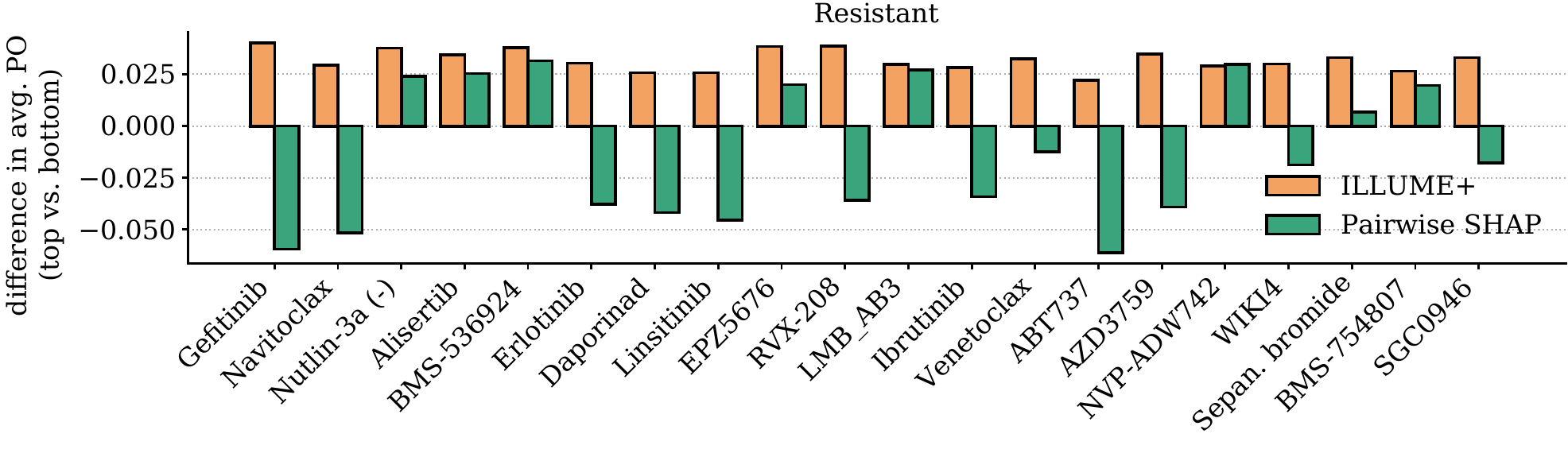}
\caption{Comparison with pairwise SHAP. Difference in mean pathway overlap (PO) between the top and bottom 1,000 ranked gene pairs across each drug. Positive values indicate greater pathway overlap among highly ranked interactions.}
\label{fig:relative_mean_variation}
\end{figure*}

\spara{Comparison with pairwise SHAP.} 
To further assess whether interaction rankings prioritize biologically coherent gene pairs, we compared \illora and pairwise SHAP in terms of pathway overlap (PO). For each drug, we computed the mean PO among the highest-ranked 1,000 interaction pairs and contrasted it with the mean PO among the lowest-ranked 1,000 pairs. 
As shown in Figure~\ref{fig:relative_mean_variation}, \illora consistently exhibits a clear separation between the two groups, with top-ranked interactions displaying greater mean pathway overlap across all drugs. In contrast, pairwise SHAP shows much weaker discrimination: for several drugs, the mean pathway overlap of bottom-ranked interactions exceeds that of top-ranked interactions. This indicates that pairwise SHAP rankings are less effective at prioritizing biologically coherent gene-gene interactions, whereas \illora more reliably concentrates high-ranking interactions among genes participating in related biological pathways.

\section{Polarity graph analysis}
\label{App:polarity graph}

In \Cref{sec:results}, we analyze a gene–gene interaction graph derived from the extracted decision rules.
In this graph, nodes correspond to genes, and edge weights quantify the strength of interaction between gene pairs, as measured by lift.
Separate gene-gene interaction graphs are constructed for the sensitive and resistant classes, reflecting class-specific interaction patterns.

To explicitly capture the class-discriminative role of gene–gene interactions, we additionally introduce the polarity graph $\graphPol = (V_\Delta, E_\Delta, w_\Delta)$.
This graph is obtained by computing, for each gene pair, the difference between its lift value in the sensitive class and its lift value in the resistant class.
As a result, edge weights encode both the magnitude and direction of class specificity, highlighting interactions that preferentially characterize sensitivity or resistance.
Positive edge weights indicate stronger interactions in the \resistant class, and negative values indicate the opposite.
Edge weights close to zero correspond to interactions that are similarly represented in both classes and are therefore weakly discriminative.
Based on this graph structure, we investigate whether genes within the same biological pathway exhibit coordinated, non-random changes in pairwise interaction strength between drug response classes (sensitive versus resistant), beyond what would be expected from generic graph structure alone.

For each MoA-related pathway \(\pi\), we compute a \emph{polarity} score \(\Delta(\pi)\) defined as the mean differential interaction weight over edges internal to the pathway, i.e.:
\[
\Delta(\pi)
=
\frac{1}{|E_{\Delta_\pi}|}
\sum_{(u,v)\in E_{\Delta_\pi}} w_{ \Delta _{uv}},
\qquad
E_{\Delta_\pi}=\{(u,v)\in E_\Delta :\ u,v\in \pi\},
\]
where \(w_{ \Delta _{uv}}=w^{(resistant)}_{uv}-w^{(sensitive)}_{uv}\) is the difference between class-conditional lift scores in the resistant and sensitive classes and \(|E_{\Delta_\pi}|\) denotes the number of 
edges with both incident nodes within $\pi$. 
We restrict the analysis to pathways with at least $5$ nodes in $V_{\Delta}$ and at least $3$ edges in $E_{\Delta_\pi}$. 
A positive \(\Delta(\pi)\) indicates that interactions among pathway genes are, on average, stronger under label \resistant than under label \sensitive, whereas negative values indicate the opposite.

To assess statistical significance, we construct a null distribution using a strength-matched randomization procedure.
First, all genes in 
$\graphPol$ are discretized into $30$ bins according to their node strength.
For a given pathway 
$\pi$, we generate null gene sets with the same cardinality as $\pi$.
Each null set is obtained by independently replacing every gene in 
$\pi$ with a gene sampled uniformly at random from the same strength bin.
This procedure preserves the node-strength profile of the pathway while removing pathway-specific structure.
The resulting collection of randomized gene sets defines the null distribution used for significance testing.

\begin{figure}[t]
    \centering
    \begin{tabular}{c|c}
    \includegraphics[width=0.49\linewidth]{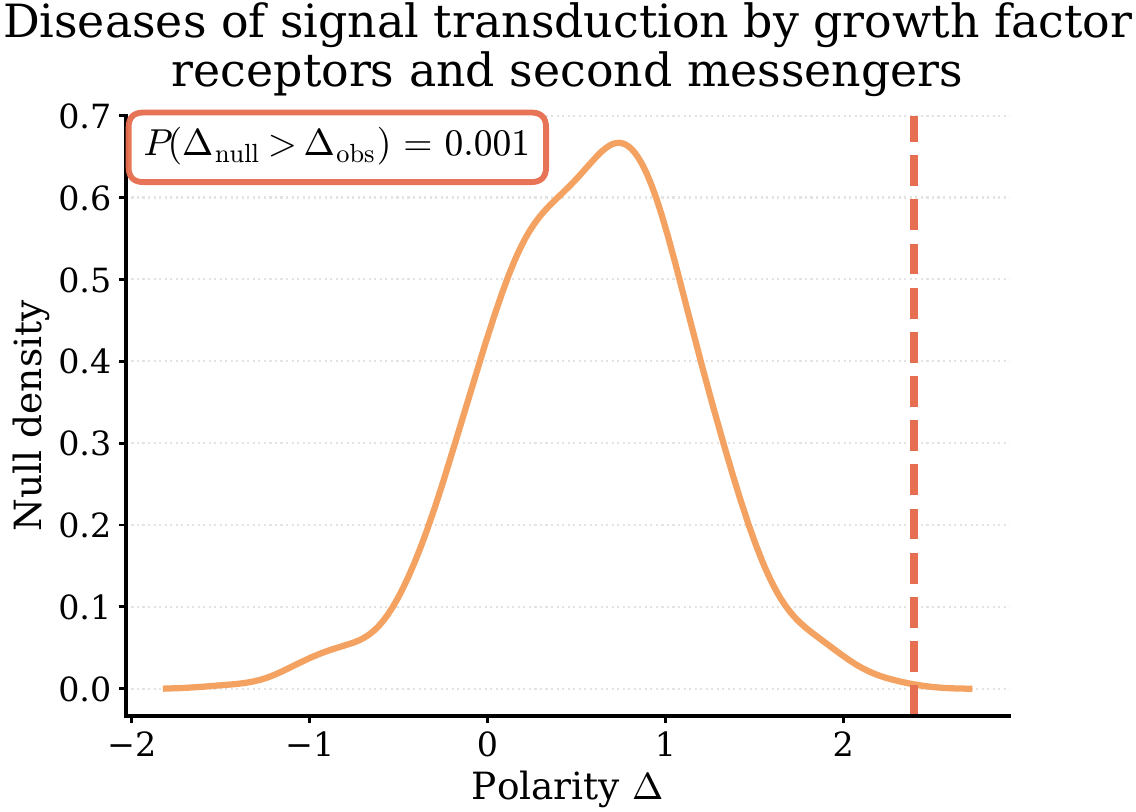}   &
    \includegraphics[width=0.46\linewidth]{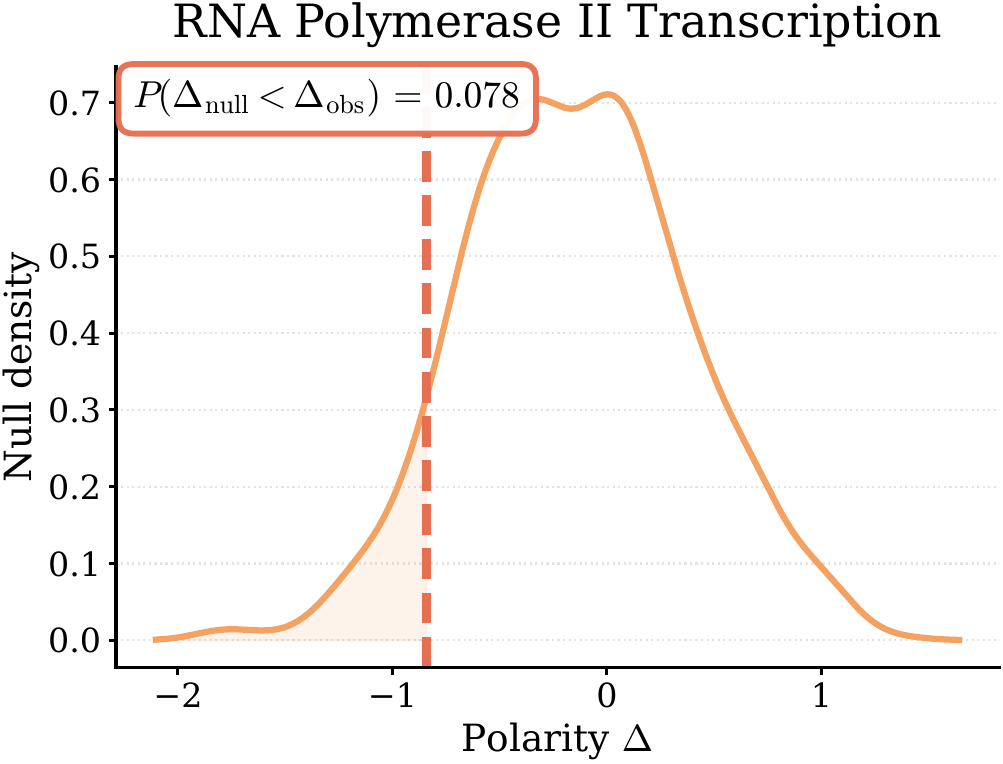}  
    \end{tabular}
    \caption{Polarity graph analysis. Null distribution of polarity and empirical polarity of the pathway specified above for drugs \drug{Gefitinib} (left) and \drug{Erlotinib} (right). }
    \label{fig:example_graphPol}
\end{figure}

The observed \(\Delta(\pi)\) is compared to the null 
distribution to obtain empirical \(p\)-values. 
Because a pathway can be associated with either increased sensitivity or increased resistance, we compute the smaller of the left- and right-tail empirical $p$-values.
We find that approximately 
$15\%$ of the
pathways attain an empirical $p$-value below  $0.1$, while $6\%$  fall below $0.05$.
Figure~\ref{fig:example_graphPol} provides an example.

\section{Validation on additional datasets}
\label{App:Additional datasets}

\spara{Additional datasets from GDSC database.}
To enable biological validation based on putative target recovery, the main analysis was restricted to the $20$ GDSC drugs for which the annotated putative targets were retained by \BORUTA during feature selection. 
This restriction is necessary for target-based metrics, but not for other evaluations.  
Therefore, to test whether the conclusions extend beyond this subset, we considered $5$ additional GDSC drugs for which \BORUTA discards the putative target. 
For these drugs, \illora remained more robust than SHAP, with median robustness of $0.626$ versus $0.497$ for the \sensitive class and $0.473$ versus $0.381$ for the \resistant class, corresponding to relative improvements of $26.1\%$ and $24.1\%$, respectively. 
Both differences were highly significant according to a Mann-Whitney U test ($p$-value$<10^{-6}$).
We then evaluated biological coherence at the pathway level, by testing whether significantly enriched pathways in the explanations were consistent with the known drug mechanism of action. 
Across these additional GDSC drugs, \illora recovered more MoA-related enriched pathways than SHAP. 
For example, for \drug{SGC0946}, a DOT1L methyltransferase inhibitor, \illora highlighted olfactory-receptor expression pathways in \sensitive cell lines, consistent with their dependence on H3K79 methylation. 
For \drug{Dasatinib}, explanations for the \resistant class were enriched for extracellular-matrix organization, collagen formation, and assembly of collagen fibrils and other multimeric structures, in line with processes associated with tyrosine-kinase activity and cancer drug resistance.

\spara{Additional datasets from PRISM database.}
To further assess whether the observed advantages of \illora generalize beyond the GDSC benchmark, we performed an additional validation experiment on PRISM~\cite{corsello2020discovering}, an independent large-scale pharmacogenomic screening resource including oncological and non-oncological compounds. In contrast to GDSC, which provides $IC_{50}$ measurements derived from drug-response curves, PRISM reports log-fold-change viability estimates measured at a single drug concentration. For this reason, we use PRISM solely as a complementary external validation setting.
In particular, we consider a sample of 14 drugs from PRISM and compared \illora against SHAP using two complementary criteria. First, we evaluated internal robustness using the same local cosine-similarity metric adopted in the main experiments. Second, we assessed biological validity by testing the Reactome pathways significantly enriched by the feature-importance rankings produced by each method, using false-discovery-rate correction. Results are aggregated across training, validation, and test splits; the same qualitative trend is observed on each split separately.

\begin{table}[t]
\centering
\caption{External validation on PRISM. We report mean explanation robustness, the mean number of significantly enriched Reactome pathways per drug, and the fraction of drugs for which at least one pathway is enriched.}
\label{tab:prism_validation}
\small
\setlength{\tabcolsep}{5pt}
\begin{tabular}{lccc}
\toprule
Method & 
\makecell{Mean\\robustness} & 
\makecell{Enriched pathways\\per drug} & 
\makecell{Drugs with\\enrichment} \\
\midrule
\illora & 0.3058 & 0.6571 & 0.4000 \\
SHAP   & 0.2510 & 0.1715 & 0.1143 \\
\bottomrule
\end{tabular}
\end{table}

\ILLoRA outperforms SHAP on all considered metrics. In particular, \ILLoRA achieves higher mean robustness, identifies a larger number of significantly enriched Reactome pathways per drug, and yields at least one enriched pathway for a larger fraction of drugs. Moreover, among the enriched pathways, \ILLoRA identifies more pathway-drug associations directly related to known mechanisms of action: 6 associations across 4 drugs, compared with 4 associations across 2 drugs for SHAP. These results support the generalizability of our conclusions beyond the GDSC setting and suggest that \ILLoRA provides more stable and biologically grounded explanations also on an independent drug-screening resource.

\end{document}